\documentclass{article}

\usepackage{microtype}
\usepackage{graphicx}
\usepackage{subfigure}
\usepackage{booktabs} 
\usepackage{url}            
\usepackage{amsfonts}       
\usepackage{nicefrac}       
\usepackage{bm}
\usepackage{amssymb}
\usepackage{amsmath}
\usepackage{amsthm}
\usepackage{array}
\usepackage{tabularx}
\usepackage{multirow}
\usepackage{verbatim}

\usepackage{color}
\newcommand{\partitle}[1]{\noindent \textbf{#1.}}
\usepackage[Symbol]{upgreek}
\usepackage{stfloats}
\usepackage{diagbox}
\usepackage{natbib}
\usepackage{dsfont}

\usepackage{hyperref}

\usepackage[accepted]{icml2020}

\icmltitlerunning{Submission and Formatting Instructions for ICML 2020}

\begin{document}

\twocolumn[
\icmltitle{Graph Convolutional Network for Recommendation \\ with Low-pass Collaborative Filters}




\begin{icmlauthorlist}
\icmlauthor{Wenhui Yu}{to}
\icmlauthor{Zheng Qin}{to,co}
\end{icmlauthorlist}

\icmlaffiliation{to}{Tsinghua University}
\icmlaffiliation{co}{Corresponding author}
\icmlcorrespondingauthor{Wenhui Yu}{yuwh16@mails.tsinghua.edu.cn}
\icmlcorrespondingauthor{Zheng Qin}{qingzh@mail.tsinghua.edu.cn}

\icmlkeywords{Graph Fourier Transform, Graph Convolution, Low-pass Graph Filter, Deep Neural Network, Collaborative Filtering, Item Recommendation.}

\vskip 0.3in
]



\printAffiliationsAndNotice{}  

\begin{abstract}
\textbf{G}raph \textbf{C}onvolutional \textbf{N}etwork (\textbf{GCN}) is widely used in graph data learning tasks such as recommendation. However, when facing a large graph, the graph convolution is very computationally expensive thus is simplified in all existing GCNs, yet is seriously impaired due to the oversimplification. To address this gap, we leverage the \textit{original graph convolution} in GCN and propose a \textbf{L}ow-pass \textbf{C}ollaborative \textbf{F}ilter (\textbf{LCF}) to make it applicable to the large graph. LCF is designed to remove the noise caused by exposure and quantization in the observed data, and it also reduces the complexity of graph convolution in an unscathed way. Experiments show that LCF improves the effectiveness and efficiency of graph convolution and our GCN outperforms existing GCNs significantly. Codes are available on \url{https://github.com/Wenhui-Yu/LCFN}.
\end{abstract}

\section{Introduction}
\label{sec:introduction}
Graph convolution \cite{graph_fourier} is the extension of convolution from the Euclidean domain to the graph domain and \textbf{G}raph \textbf{C}onvolutional \textbf{N}etwork (\textbf{GCN}) is the extension of \textbf{C}onvolutional \textbf{N}eural \textbf{N}etwork (\textbf{CNN}). GCN is widely explored in graph learning tasks such as recommendation since it can extract abundant information from the graph data. However, the eigen-decomposition of a large matrix is required in the graph convolution \cite{graph_fourier}, which is extremely space- and time-consuming in real-world applications. To avoid the eigen-decomposition, all existing GCNs constrain (even fix) the convolutional kernel and simplify graph convolution to graph propagation \cite{GCN,semi_super}. \textbf{G}raph \textbf{N}eural \textbf{N}etwork (\textbf{GNN}) propagates signals through the graph, and refines the signal of each node with its neighborhoods \cite{GNN_zongshu}. By constraining the kernel, GCNs lose the ability of convolution and degenerate to GNNs (discussed in detail in Section \ref{subsec:comparison}). To address the oversimplification issue, we propose a new GCN that can really leverage the ability of graph convolution and validate the effectiveness in the recommendation task. In our GCN, we propose a \textbf{L}ow-pass \textbf{C}ollaborative \textbf{F}ilter (\textbf{LCF}) both to filter the noise in observed data and to reduce the time consumption of graph convolution without hurting its ability.

\begin{figure*}[ht]
    \centering
    \includegraphics[scale = 0.38]{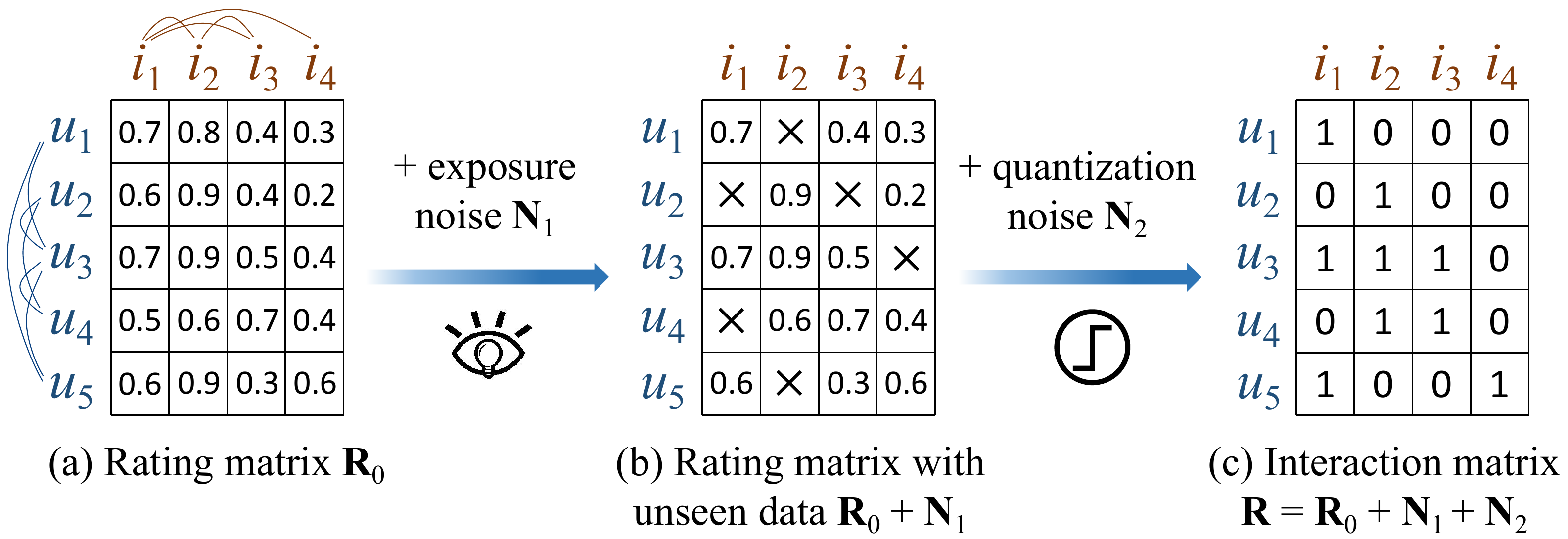}
    \caption{The interaction matrix and different kinds of noise. Curves in Figure \ref{fig:interaction_matrix}(a) indicate a user graph and an item graph. Two users are connected if they have interacted with the same item, and two items are connected if they have been interacted by the same user.}
    \label{fig:interaction_matrix}
\end{figure*}

LCF is devised to remove the noise in the observed data. In this paper, we adopt implicit feedback data (interactions such as ``purchase'', ``click'', ``watch'' records, denoted by binary variables) since it is generally applicable in recommendation tasks. A matrix for observed interactions is illustrated in Figure \ref{fig:interaction_matrix}(c), and Figure \ref{fig:interaction_matrix} shows how it is generated: There is an unknown underlying numeric rating matrix ${\bm{{\rm R}}}_0$ indicating the true preference score for each user towards each item, shown in Figure \ref{fig:interaction_matrix}(a). When browsing items on a website, users can only browse a small proportion of the mass items, thus many items are not exposed to the users, including many preferred items. Reflected in the matrix representation, many elements in ${\bm{{\rm R}}}_0$ are set to 0 randomly, shown in Figure \ref{fig:interaction_matrix}(b). To simulate this process, we add exposure noise ${\bm{{\rm N}}}_1$ to the rating matrix ${\bm{{\rm R}}}_0$. For the remaining exposed items, a user will interact with them only if she like them. In the matrix representation, we discretize the matrix ${\bm{{\rm R}}}_0+{\bm{{\rm N}}}_1$ with a threshold, which is set as 0.5 in this example. 
The quantization noise is denoted as ${\bm{{\rm N}}}_2$, and we have ${\bm{{\rm R}}} = {\bm{{\rm R}}}_0+{\bm{{\rm N}}}_1+{\bm{{\rm N}}}_2$. Our task is to reconstruct ${\bm{{\rm R}}}_0$ from observed ${\bm{{\rm R}}}$ and rank all items for each user based on the reconstruction. 

Considering ${\bm{{\rm R}}}_0$ is low frequency due to the strong dependence and ${\bm{{\rm N}}}_1$ and ${\bm{{\rm N}}}_2$ are high frequency due to the randomness (we will introduce the reason in detail in Section \ref{subsec:fourier_trans}), we can remove noise by low-pass filtering. However, different from the images, ${\bm{{\rm R}}}$, ${\bm{{\rm R}}}_0$, ${\bm{{\rm N}}}_1$ and ${\bm{{\rm N}}}_2$ are all 2D graph signals, thus conventional filters are not effective. For signals defined in the Euclidean domain like images and timing signals, the dependence is based on position, such as pixels near in the space position or sampling points near in the time axis. Nevertheless, for signals defined in the graph domain, the dependence is based on graph connections, for example, we consider the dependence of two rows/columns of ${\bm{{\rm R}}}_0$ based on the user/item graph shown in Figure \ref{fig:interaction_matrix}(a). To remove the graph noise, we devise a 2D graph filter called LCF and reconstruct ${\bm{{\rm R}}}_0$ by it: $\hat{\bm{{\rm R}}}_0 = LCF({\bm{{\rm R}}})$, where $\hat{\bm{{\rm R}}}_0$ is the estimation. Similar to the filter, we also extend 2D convolution from the Euclidean domain to the graph domain to propose GCN. We then inject LCF into GCN to propose our \textbf{LCF} \textbf{N}etwork (\textbf{LCFN}). LCF not only improves the accuracy by removing the noise, but also reduces the computation of graph convolution since only a small proportion of eigen-vectors are required. To the best of our knowledge, LCFN is the first GCN to use the original graph convolution.

To summarize, this work makes the following contributions:

\begin{itemize}
\vspace{-2mm}
{\item We introduce the 2D graph Fourier transform, and then design the graph filter LCF and graph convolution based on it.}
\vspace{-1mm}
{\item We design a novel GCN named LCFN for recommendation with LCF to improve the performance by removing noise and to improve the efficiency of graph convolution.}
\vspace{-1mm}
{\item We devise comprehensive experiments on two real-world datasets to demonstrate the effectiveness and efficiency of LCFN.}
\end{itemize}
\vspace{-2mm}

\section{Related Work}
\label{sec:related_work}
After the collaborative filtering (CF) model was proposed \cite{item-based,user-based}, recommender systems have boomed in decades. Among various CF methods, matrix factorization (MF) \cite{MF,BPR}, which encodes the preference of users by underlying embeddings (i.e., latent factors), is a basic yet the most effective recommender model. To improve the presentation capability, many variants have been proposed \cite{NCF,NGCF}. Among these models, neural network-based models and graph-based models are the most relevant to our work.

\partitle{Neural Network-based Recommendation} These years, Neural Network achieves significant success and is widely used in various machine learning and information retrieval domains, including recommendation. \citet{wide_deep,NCF} utilized wide and deep networks to learn the nonlinear combination of embeddings. \citet{attentive} explored the attention mechanism to highlight important elements. \citet{VBPR,AES} explored side information to enhance the performance of recommendation. \citet{seque_rec1,seque_rec2} regarded recommendation as a sequence prediction task and utilized RNN or CNN to deal with this issue. Though extensively explored, these models are based on conventional networks and omit the graph structure.

\partitle{Graph-based Recommendation} As a link prediction task of a bipartite graph, graph tools are widely used in the recommendation domain. \citet{hyper_nips} used graph regularization to smooth the embeddings of connected nodes (users and items in recommendation context). \citet{SPLR} extracted spectral features for nodes from the hypergraph Laplacian matrices to optimize model learning. However, these shallow models only get very limited information. \citet{NGCF,GNN_WebScale,GNN_MatCom} utilized deep GNNs for recommendation tasks. GNNs propagate embeddings through the graph to ensure that connected nodes have similar latent representations. 
Though GNNs perform well, there is still a large room for improvement --- GNNs only smooth the embeddings while cannot extract high-level patterns for the graph signal. To address this gap, GCNs are proposed.

\citet{graph_fourier} proposed graph convolution to extract the dependence pattern from the graph data. Since graph convolution on large graphs is extremely time-consuming, \citet{GCN,semi_super} first simplified it by various approximations, unfortunately, degenerate GCNs to GNNs. By following them, existing GCNs are indeed GNNs, or variants of GNNs \cite{GCN_rec1,GCN_rec2,SCF}. Also, as a result, many GNNs are claimed as GCNs \cite{GNN_MatCom,GNN_WebScale} (We follow the definition of GNN and GCN in \cite{GNN_zongshu}). In this paper, we devise LCF to reduce the computation of graph convolution and to remove the noise. Different from existing GCNs, our simplification strategy enhances the graph convolution rather than degenerating it.

\section{Preliminaries}
\label{sec:preliminaries}
In this section, we introduce some background knowledge of the signal processing in the Euclidean domain. 

\subsection{Fourier Transform}
\label{subsec:fourier_trans}
Fourier transform is a widely-used tool for frequency domain analysis. It converts the signal from the time domain to the frequency domain. For a continuous signal $f(t)$, the Fourier transform is: $F(\omega)=\int_\mathbb{R} f(t)\text{e}^{-j \omega t}dt$ and the inverse transform is $f(t)=\frac{1}{2\pi}\int_\mathbb{R} F(\omega)\text{e}^{ j \omega t}d\omega$. For a digital signal (discrete signal) denoted as $\{s_n\}$, the Fourier transform is: $S_k=\sum_{n=0}^{N-1}(s_n)\text{e}^{-\frac{2\pi j}{N}kn}$ and the inverse transform is $s_n=\frac{1}{N}\sum_{k=0}^{N-1}(S_k)\text{e}^{\frac{2\pi j}{N}kn}$, where $j$ is the imaginary unit and $N$ is the length of the series $\{s_n\}$. 

\begin{figure}[ht]
    \centering
    \subfigure[$\!s_n^1\!=\!\sin(\frac{\pi}{10}n)$]{
        \includegraphics[scale = 0.36]{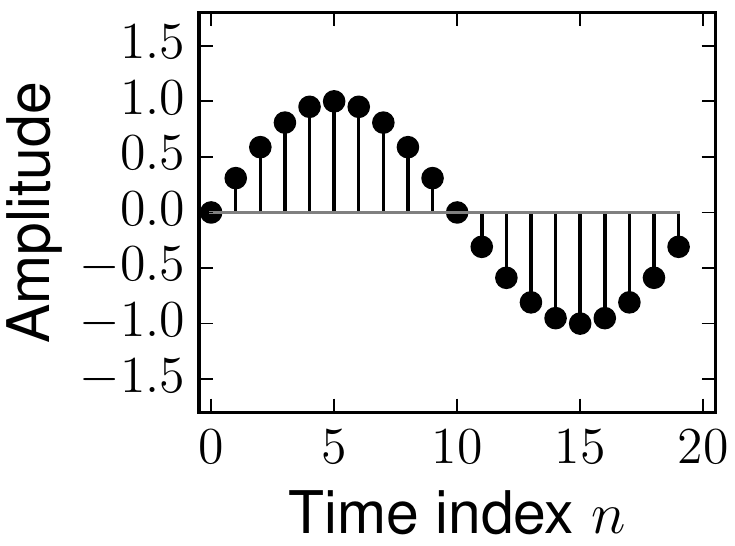} 
        \label{subfig:s1}
    }
    \hspace{-3.5mm}
    \subfigure[$\!s_n^2\!=\!\sin(\frac{7\pi}{10}n)$]{
        \includegraphics[scale = 0.36]{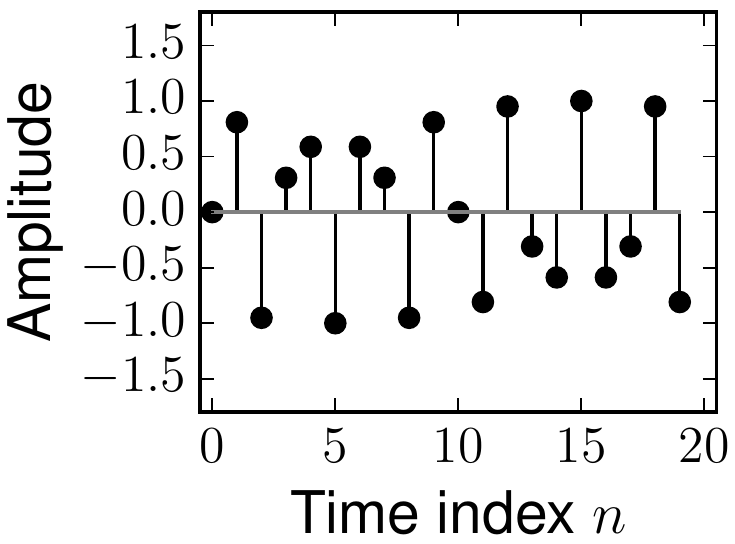}
        \label{subfig:s2}
    }
    \hspace{-3.5mm}
    \subfigure[$\!s_n^3\!=\!s_n^1\!+\!s_n^2$]{
        \includegraphics[scale = 0.36]{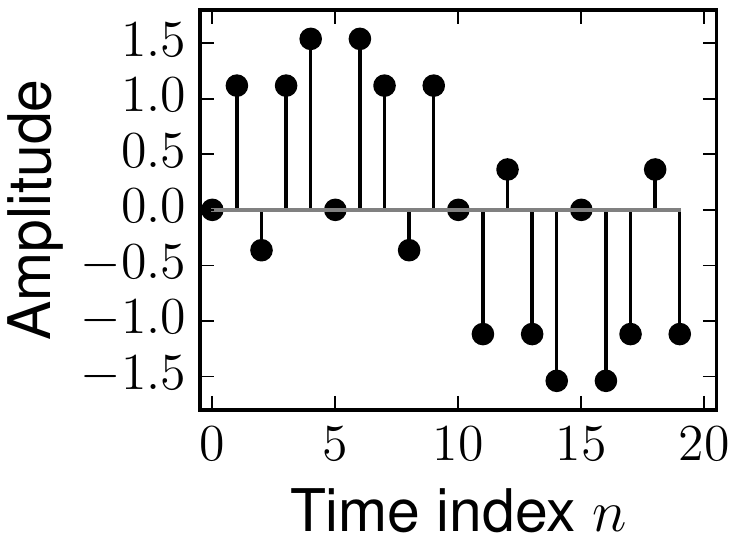}
        \label{subfig:s3}
    }
    
    \subfigure[$\!S_k^1\!=\!|\mathcal{F}(s_n^1)|$]{
        \includegraphics[scale = 0.39]{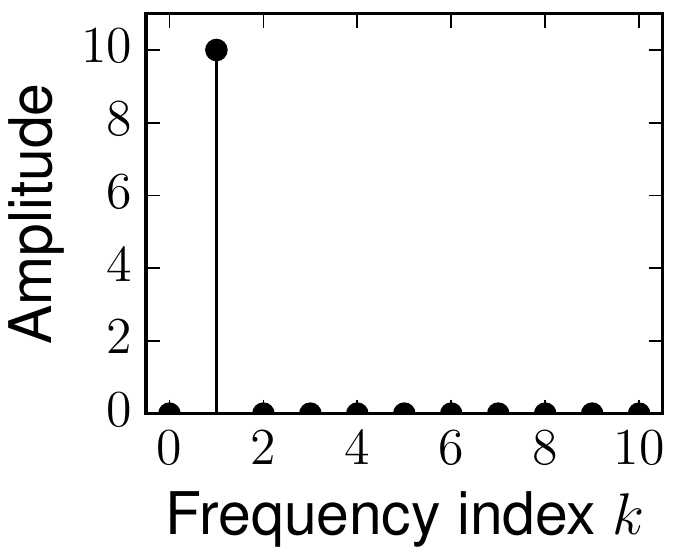}
        \label{subfig:f1}
    }
    \hspace{-3.5mm}
    \subfigure[$\!S_k^2\!=\!|\mathcal{F}(s_n^2)|$]{
        \includegraphics[scale = 0.39]{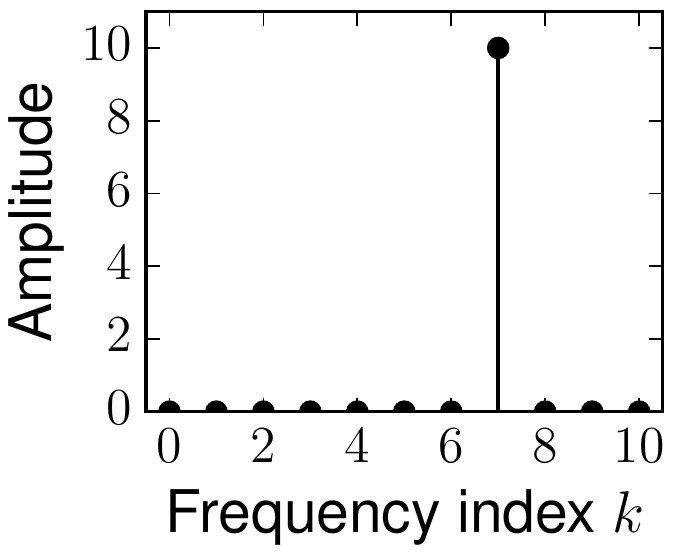}
        \label{subfig:f2}
    }
    \hspace{-3.5mm}
    \subfigure[$\!S_k^3\!=\!|\mathcal{F}(s_n^3)|$]{
        \includegraphics[scale = 0.39]{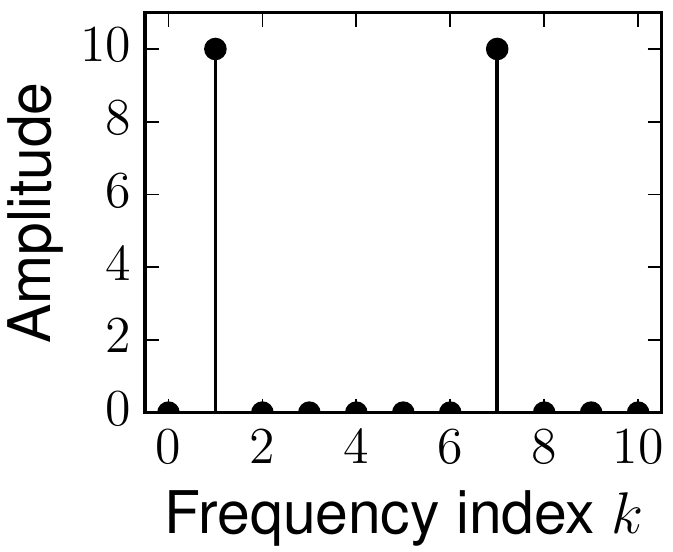}
        \label{subfig:f3}
    }
    \caption{Signals with different frequencies in the time domain and frequency domain.}
    \label{fig:signals}
\end{figure}

In this paper, we denote the Fourier transform as $\mathcal{F}(\;)$ and the inverse transform as $\mathcal{F}^{-1}(\;)$. Figure \ref{fig:signals} shows some signals with different frequencies. Figures \ref{fig:signals}(a)(b)(c) show signals in the time domain and Figures \ref{fig:signals}(d)(e)(f) show signals in the frequency domain. $s_n^1$ in Figure \ref{subfig:s1} is a low-frequency signal. In the time domain, $s_n^1$ is very smooth and shows strong dependence between two adjacent sampling points. In the frequency domain, the signal is distributed on the low frequency (shown in Figure \ref{subfig:f1}). In contrast, $s_n^2$ in Figure \ref{subfig:s2} is a high-frequency signal thus shows weak dependence and is distributed on the high frequency (shown in Figure \ref{subfig:f2}). $s_n^3$ in Figure \ref{subfig:s3} is the addition of $s_n^1$ and $s_n^2$. They are fully mixed in the time domain, nevertheless can be easily separated in the frequency domain (shown in Figure \ref{subfig:f3}). In telecommunication, low-frequency signals are usually contaminated by high-frequency noise when transmitting through the channel, and we can reconstruct the signal by a low-pass filter. 

As discussed, the frequency in frequency domain reflects the dependence in time domain, however, in graph data, the dependence is based on node connections. Recalling the example shown in Figure \ref{fig:interaction_matrix}, $u_1$ and $u_3$ are connected users, the two rows for them in $\bm{{{\rm R}}}_0$ are $(0.7,0.8,0.4,0.3)$ and $(0.7,0.9,0.5,0.4)$, and in $\bm{{{\rm N}}}_1$ are $(0,-0.8,0,0)$ and $(0,0,0,-0.4)$. We can see that the true preference of two connected nodes shows strong dependence thus $\bm{{{\rm R}}}_0$ is low-frequency, nevertheless the noise shows weak dependence due to the random position, thus is high-frequency. We can remove the high-frequency graph noise from the low-frequency graph signal by a low-pass graph filter.

\subsection{Convolution}
\label{subsec:convolution}
In addition to Fourier transform, convolution is a very important signal processing tool which is widely used in CNNs. For two continuous signals $f(t)$ and $g(t)$, the convolution of them is defined as: $f(t)\!*\!g(t)\!=\!\int_\mathbb{R}f(\tau)g(t-\tau) d\tau$. For two digital signals $\{s_n\}_{n=0\cdots N-1}$ and $\{k_m\}_{m=0\cdots M-1}$, the convolution is: $s'_n=\sum_{m=0}^{M-1}k_m(s_{n-m})$, where $N$ and $M$ indicate the length of the signals. In machine learning tasks, $M$ is usually much smaller than $N$, and $\{k_m\}$ is called the convolutional kernel, $\{s'_n\}$ is the feature map. We can extract local dependence information from the signal by convolution. Intuitively speaking, if $\{s_n\}$ is very similar to $\{k_m\}$ in a local area, $\{s'_n\}$ will be very large in this position.

We also introduce a widely-used property of convolution, which is crucial in graph convolution design: \textit{Convolution Theorem} \cite{conv_theo} tells us that convolution in the time domain is equivalent to the product in the frequency domain, i.e., $f(t) \!*\!g(t)\!\! =\!\! \mathcal{F}^{-1} (\mathcal{F} (f(t)) \mathcal{F} (g(t)))$. For graph data, convolution is hard to define directly due to the irregular structure of graphs, thus is defined in this way.

\section{2D Low-pass Graph Convolution}
\label{sec:2d_graph_transform}
In this paper, bold uppercase letters refer to matrices. Assuming there are $M$ users and $N$ items in total, we use matrix $\bm{{{\rm R}}} \in \mathbb{R}^{M \times N}$ to denote the interactions between users and items. $\bm{{{\rm R}}}_{ui} = 1$ if user $u$ has interacted with item $i$ and $\bm{{{\rm R}}}_{ui} = 0$ otherwise. Our task is to predict the missing entities, i.e., 0 in $\bm{{{\rm R}}}$.

\subsection{2D Graph Fourier Transform}
\label{subsec:graph_FT}
To propose LCF and the graph convolution, we need to propose graph Fourier transform first. \citet{graph_fourier} extended Fourier transform from the Euclidean domain to the graph domain. In this section, we follow \citet{graph_fourier} and propose a 2D graph Fourier transform. For a continuous 2D signal in Euclidean domain $f(x,y)$, the Fourier transform is:
\begin{small}
	\begin{eqnarray}
		\left.\begin{aligned}
		\label{equ:fourier_transform}
		F\!\left(\phi,\!\psi \right)\!\!=\!\!\!\underset{\mathbb{R}}{\mathop \int }\!\!\underset{\mathbb{R}}{\mathop \int }\!f\!\left( x,\!y \right)\!{\text{e}^{\!-\!j \left(\!\phi x+ \psi y \!\right)}}dxdy \!\!=\!\! \left\langle\! f\!\left(x,\!y\right)\!,\text{e}^{\!-j\phi x}\!\!\!\cdot\!\text{e}^{\!-j\psi y} \!\right\rangle,
		\end{aligned}
		\right.
	\end{eqnarray}
\end{small}which is defined as the inner product of the signal and the product of transform bases on two dimensions. Taking dimension $x$ as an example, the transform base $\text{e}^{- j\phi x}$ and corresponding frequency $\phi$ are defined as the eigen-function $\text{e}^{- j\phi x}$ and eigen-value $-j\phi$ of the differential operator: $\nabla\!_x\text{e}^{- j\phi x} \!=\! -j\phi \text{e}^{- j\phi x}$.

To extend the above definition to the graph domain, we utilize the differential operator for graph signals: Laplacian matrices. We construct Laplacian matrices $\bm{{{\rm L}}}^U$ and $\bm{{{\rm L}}}^I$ for user and item graphs. The transform bases and the frequencies for graph Fourier transform are defined as the eigen-vectors and eigen-values of the Laplacian matrices. 

Now, we introduce how to construct the Laplacian matrices. Considering hypergraphs are more powerful in representation and provide more information than simple graphs, we follow \citet{SPLR} to use two hypergraphs instead of simple graphs shown in Figure \ref{fig:interaction_matrix}(a). The hypergraph is a generalization of the simple graph, where an edge connects any number of nodes rather than just two, called hyperedge. Assuming there are $M$ nodes and $N$ hyperedges, a hypergraph is usually represented with the incidence matrix $\bm{{{\rm H}}}\in \mathbb{R}^{M\times N}$. Each row of $\bm{{{\rm H}}}$ is for a node and each column is for a hyperedge. $\bm{{{\rm H}}}_{ij}=1$ if node $i$ is connected by hyperedge $j$ and $\bm{{{\rm H}}}_{ij}=0$ otherwise. The Laplacian matrix $\bm{{{\rm L}}}$ is defined as \cite{hyper_laplacian}:
\begin{small}
	\begin{eqnarray}
	\left.\begin{aligned}
	\label{equ:Laplacian}
	\bm{{{\rm L}}} =  {\bm{{\rm I}}} -{\bm{{\rm D}}}^{-\frac{1}{2}} {\bm{{\rm H}}} {\bm{{\rm W}}} {\bm{{\rm \Delta}}}^{-1} {\bm{{\rm H}}}^\mathsf{T} {\bm{{\rm D}}}^{-\frac{1}{2}},
	\end{aligned}
	\right.
	\end{eqnarray}
\end{small}where the diagonal matrix $\bm{{{\rm D}}}\in \mathbb{R}^{M \times M}$ denotes degrees of nodes, the diagonal matrix $\bm{{{\rm W}}} \in \mathbb{R}^{N \times N}$ denotes weights of hyperedges, and the diagonal matrix $\bm{{{\rm \Delta}}}\in \mathbb{R}^{N \times N}$ denotes the degrees of hyperedges. The effect of the Laplacian matrix is to take the difference for a signal on the hypergraph \cite{hyper_laplacian,SPLR}. 

As hypergraphs can represent the interaction data naturally, we proposed two hypergraphs for users and items. In the user hypergraph, users are nodes and items are hyperedges and in the item hypergraph, items are nodes and users are hyperedges. Laplacian matrices for users ${\bm{{\rm L}}}^U$ and for items ${\bm{{\rm L}}}^I$ can be constructed by setting ${\bm{{\rm H}}}={\bm{{\rm R}}}$ and ${\bm{{\rm H}}}={\bm{{\rm R}}}^\mathsf{T}$ in Equation (\ref{equ:Laplacian}), respectively. After getting the differential operators ${\bm{{\rm L}}}^U$ and ${\bm{{\rm L}}}^I$, we decompose them by eigen-decomposition. Considering ${\bm{{\rm L}}}^U$ and ${\bm{{\rm L}}}^I$ are positive semi-definite, we have ${\bm{{\rm L}}}^U={\bm{{\rm P}}}{\bm{{\rm \Lambda}}}{\bm{{\rm P}}}^\mathsf{T}$ and ${\bm{{\rm L}}}^I={\bm{{\rm Q}}}{\bm{{\rm \Sigma}}}{\bm{{\rm Q}}}^\mathsf{T}$. Eigen-vector matrices ${\bm{{\rm P}}}$ and ${\bm{{\rm Q}}}$ are graph Fourier transform bases for user and item dimensions, respectively and eigen-value matrices ${\bm{{\rm \Lambda}}}\!=\!diag(\lambda_1,\lambda_2,\cdots,\lambda_M)$ and ${\bm{{\rm \Sigma}}}\!\!=\!\!diag(\sigma_1,\sigma_2,\cdots,\sigma_N)$ are corresponding frequencies (all eigen-values are in ascending order). Similar to Equation (\ref{equ:fourier_transform}), the graph Fourier transform of signal ${\bm{{\rm R}}}$ is the inner product of ${\bm{{\rm R}}}$ and the product of transform bases on two dimensions:
\begin{small}
\begin{align}
\label{equ:graph_fourier}
\tilde{\bm{{\rm R}}}_{\phi\psi} =\left\langle {\bm{{\rm R}}},{\bm{{\rm P}}}_{*\phi}\cdot{\bm{{\rm Q}}}_{*\psi}^\mathsf{T} \right\rangle = \sum_{u=1}^M \sum_{i=1}^N {\bm{{\rm R}}}_{ui} {\bm{{\rm P}}}_{u\phi } {\bm{{\rm Q}}}_{i\psi}.
\end{align}
\end{small}Then we have $\tilde{\bm{{\rm R}}}=\mathcal{F}_g({\bm{{\rm R}}})={\bm{{\rm P}}}^\mathsf{T} {\bm{{\rm R}}}{\bm{{\rm Q}}}$, where $\tilde{\bm{{\rm R}}}$ is the graph Fourier transform of ${\bm{{\rm R}}}$. The inverse transform is $\mathcal{F}_g^{-1}(\tilde{\bm{{\rm R}}})={\bm{{\rm P}}}\tilde{\bm{{\rm R}}}{\bm{{\rm Q}}}^\mathsf{T}$. In graph Fourier transform, the ``time domain'' corresponds to the ``graph domain'' and the ``frequency domain'' (also called ``spectral domain'') is defined by eigen-values: $
\left[
\begin{smallmatrix}
(\lambda_1,\sigma_1)  & \cdots   & (\lambda_1,\sigma_N) \\
\vdots  & \ddots   & \vdots \\
(\lambda_M,\sigma_1)  & \cdots\  & (\lambda_M,\sigma_N) \\
\end{smallmatrix}
\right]
$. Note that $\tilde{\bm{{\rm R}}}_{\phi\psi}$ in Equation (\ref{equ:graph_fourier}) is the component on frequencies $(\lambda_{\phi},\sigma_{\psi})$, thus we can also denote $\tilde{\bm{{\rm R}}}_{\phi\psi}$ as $\tilde{\bm{{\rm R}}}(\lambda_{\phi},\sigma_{\psi})$.

\begin{figure}[ht]
    \centering
    \subfigure[Discrete Euclidean domain]{
        \includegraphics[scale = 0.35]{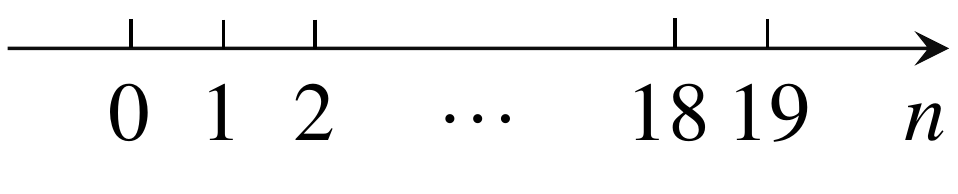}
        \label{subfig:Euclidean_domain}
    }
	\hspace{4mm}
    \subfigure[Graph domain]{
        \includegraphics[scale = 0.35]{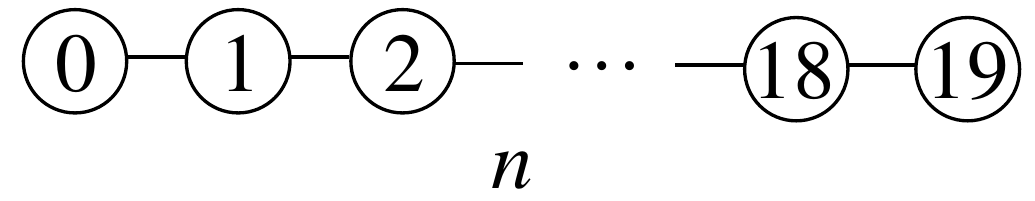}
        \label{subfig:graph_domain}
    }
    \caption{Discrete Euclidean domain can be regarded as a special case of graph domain.}
    \label{fig:domains}
\end{figure}

Now, we give a brief example to illustrate the effectiveness of the graph convolution. We recall the example introduced in Figure \ref{fig:signals}. These signals are distributed on discrete Euclidean domain, which can be regarded as a special case of graph domain (please see Figure \ref{fig:domains}).
\begin{figure}[ht]
    \centering
    \subfigure[$S_k^1=|\mathcal{F}_g(s_n^1)|$]{
        \includegraphics[scale = 0.38]{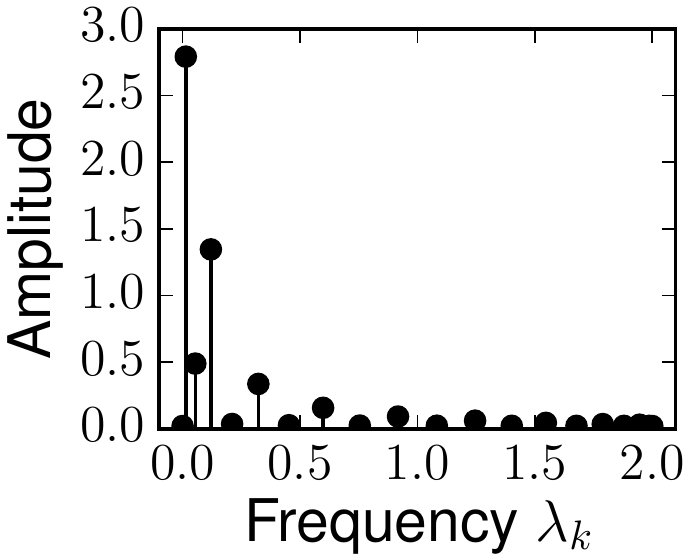} 
        \label{subfig:g_f1}
    }
    \hspace{-3.5mm}
    \subfigure[$S_k^2=|\mathcal{F}_g(s_n^2)|$]{
        \includegraphics[scale = 0.38]{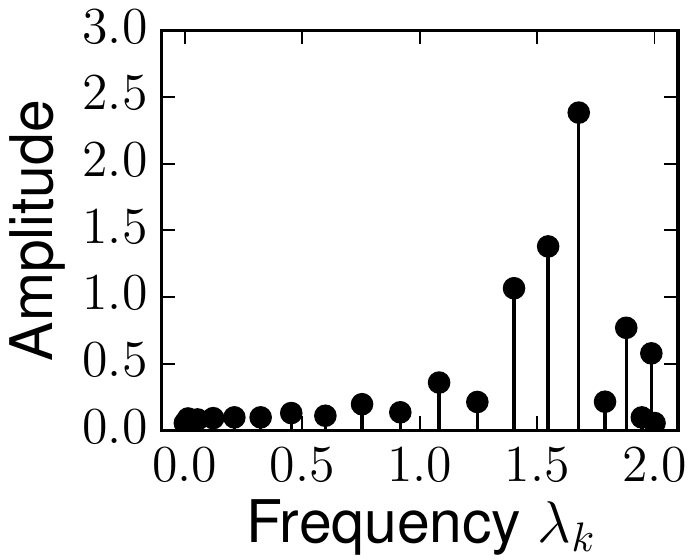}
        \label{subfig:g_f2}
    }
	\hspace{-3.5mm}
	\subfigure[$S_k^3=|\mathcal{F}_g(s_n^3)|$]{
		\includegraphics[scale = 0.38]{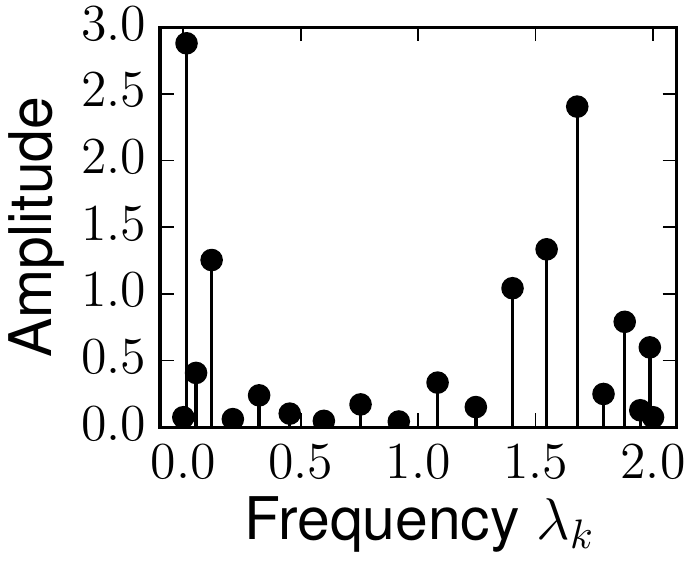}
		\label{subfig:g_f3}
	}
    \caption{Examples of graph Fourier transform.}\vspace{-2mm}
    \label{fig:graph_signal}
\end{figure}

Figure \ref{fig:graph_signal} show the graph Fourier transform of $\{s_n^1\}$, $\{s_n^2\}$, and $\{s_n^3\}$ introduced in Figure \ref{fig:signals}. We first convert $\{s_n^1\}$, $\{s_n^2\}$, and $\{s_n^3\}$ to graph signals by distributing them on the graph shown in Figure \ref{subfig:graph_domain}, and then transform them. To give a concise illustration, we leverage 1D signals, 1D transform, and simple graph in this example \cite{graph_fourier}. The abscissa in Figure \ref{fig:graph_signal} is the frequency given by eigen-values of the Laplacian matrix. We can see that graph Fourier transform can be used for frequency analysis for graph signals. 

\subsection{Graph Convolution with Low-pass Filters}
\label{subsec:low-pass_convolution}
\textbf{LCF.} For the signal ${\bm{{\rm R}}}$ in the frequency domain $\mathcal{F}_g({\bm{{\rm R}}})$, we want LCF to attenuate the high-frequency component of ${\bm{{\rm R}}}$ while retain the low-frequency component, thus we leverage the gate function $\tilde{\bm{{\rm F}}} = \left[\begin{matrix} \bm{1} & \bm{0} \\ \bm{0} & \bm{0} \end{matrix}\right]$, where $\tilde{\bm{{\rm F}}}\in \{0,1\}^{M\times N}$, $\bm{1}\in \{1\}^{\varPhi\times \varPsi}$ is an all-one matrix and $\bm{0}$ is an all-zero matrix. The passband cut-off frequencies on the user dimension and item dimension are $\lambda_{\varPhi}$ and $\sigma_{\varPsi}$, respectively. The passed signal is $LCF({\bm{{\rm R}}})=\mathcal{F}_g^{-1}(\mathcal{F}_g({\bm{{\rm R}}})\odot\tilde{\bm{{\rm F}}})$, where $\odot$ is the element-wise product. By tuning $\varPhi$ and $\varPsi$, we can remove $\bm{{{\rm N}}}_1$ and $\bm{{{\rm N}}}_2$ and retain $\bm{{{\rm R}}}_0$ as much as possible. An intuitive example is represented in Appendix \ref{app:example_GFT}.

\textbf{Graph convolution.} We extend convolution theorem to the graph domain: Given a convolutional kernel ${\bm{{\rm K}}}\in \mathbb{R}^{M\times N}$, the convolution of the signal (feature map) is defined as ${\bm{{\rm R}}}'={\bm{{\rm R}}}*_g{\bm{{\rm K}}}=\mathcal{F}_g^{-1}(\mathcal{F}_g({\bm{{\rm R}}})\odot \tilde{\bm{{\rm K}}})$, where $*_g$ is the graph convolution and $\tilde{\bm{{\rm K}}}=\mathcal{F}_g({\bm{{\rm K}}})$ is the kernel in the frequency domain. 

\textbf{Low-pass graph convolution.} We first filter the signal ${\bm{{\rm R}}}$ by LCF $\tilde{\bm{{\rm F}}}$ and then perform the convolution by a kernel $\tilde{\bm{{\rm K}}}$: ${\bm{{\rm R}}}'\!=\!\mathcal{F}_g^{-1}\! (\mathcal{F}_g({\bm{{\rm R}}})\! \odot \!\tilde{\bm{{\rm F}}} \! \odot\! \tilde{\bm{{\rm K}}})\!=\!{\bm{{\rm P}}}(({\bm{{\rm P}}}^{\mathsf{T}}{\bm{{\rm R}}}{\bm{{\rm Q}}})\!\odot\!\tilde{\bm{{\rm F}}} \!\odot\! \tilde{\bm{{\rm K}}}){\bm{{\rm Q}}}^{\mathsf{T}}$, we have:
\begin{small}
\begin{align}
{\bm{{\rm R}}}'_{u'i'}&=\sum_{\phi=1}^M\sum_{\psi=1}^N\left(\sum_{u=1}^M\sum_{i=1}^N {\bm{{\rm R}}}_{ui}{\bm{{\rm P}}}_{u\phi}{\bm{{\rm Q}}}_{i\psi}\right)\tilde{\bm{{\rm F}}}_{\phi\psi}\tilde{\bm{{\rm K}}}_{\phi\psi}{\bm{{\rm P}}}_{u'\phi}{\bm{{\rm Q}}}_{i'\psi}\nonumber\\
&=\sum_{\phi=1}^\varPhi\sum_{\psi=1}^\varPsi\left(\sum_{u=1}^M\sum_{i=1}^N {\bm{{\rm R}}}_{ui}{\bm{{\rm P}}}_{u\phi}{\bm{{\rm Q}}}_{i\psi}\right) \tilde{\bm{{\rm K}}}_{\phi\psi}{\bm{{\rm P}}}_{u'\phi}{\bm{{\rm Q}}}_{i'\psi}. \nonumber
\end{align}
\end{small}We can see that only the first $\varPhi$/$\varPsi$ eigen-vectors of ${\bm{{\rm L}}}^U$/${\bm{{\rm L}}}^V$ are needed. We define $\bar{\bm{{\rm P}}}={\bm{{\rm P}}}_{*,1:\varPhi}$, $\bar{\bm{{\rm Q}}}={\bm{{\rm Q}}}_{*,1:\varPsi}$, and $\bar{\bm{{\rm K}}}=\tilde{\bm{{\rm K}}}_{1:\varPhi,1:\varPsi}$, thus ${\bm{{\rm R}}}'=\bar{\bm{{\rm P}}}((\bar{\bm{{\rm P}}}^{\mathsf{T}}{\bm{{\rm R}}}\bar{\bm{{\rm Q}}}) \odot \bar{\bm{{\rm K}}}) \bar{\bm{{\rm Q}}}^{\mathsf{T}}$.

\section{Low-pass Collaborative Filter Network}
\label{sec:lcfn}
In this section, we propose LCFN and analyze it by comparing with some widely-used GNNs and GCNs.

\subsection{Model Structure}
LCFN is a GCN with our proposed low-pass graph convolution. The structure is illustrated in Appendix \ref{app:illustration}. 
We stack multiple convolutional layers to extract high-level features. Details of the convolutional layer are introduced below.

\partitle{Low-pass convolution}  
In LCFN, we input the feature map from the previous layer ${\bm{{\rm R}}}'$ to the current layer. Considering these feature maps are large and dense,
we represent them in a form of MF: ${\bm{{\rm R}}}' \approx {\bm{{\rm U}}}{\bm{{\rm V}}}^\mathsf{T}$, where ${\bm{{\rm U}}}\in \mathbb{R}^{M\times K}$ and ${\bm{{\rm V}}}\in \mathbb{R}^{N\times K}$ are embeddings. To avoid overfitting, we reduce the number of parameters by redefining the convolutional kernel as a rank-1 matrix $\bar{\bm{{\rm K}}} = {\bm{{\rm k}}}^U{{\bm{{\rm k}}}^V}^\mathsf{T}$, where column vectors ${\bm{{\rm k}}}^U\in \mathbb{R}^{\varPhi}$ and ${\bm{{\rm k}}}^V\in \mathbb{R}^{\varPsi}$ are kernels for user and item dimensions, respectively. In this case, performing convolution on the feature map ${\bm{{\rm R}}}'$ equals to performing convolution on the embeddings: ${\bm{{\rm U}}}' = \bar{\bm{{\rm P}}}diag({\bm{{\rm k}}}^U)\bar{\bm{{\rm P}}}^\mathsf{T}{\bm{{\rm U}}}$ and ${\bm{{\rm V}}}' = \bar{\bm{{\rm Q}}}diag({\bm{{\rm k}}}^V)\bar{\bm{{\rm Q}}}^\mathsf{T}{\bm{{\rm V}}}$.

\partitle{Transformation} We follow existing GNNs and GCNs \cite{GNN_zongshu} to use a matrix ${\bm{{\rm T}}}\in \mathbb{R}^{K\times K}$ to transform (not graph Fourier transform) the convolution of embeddings ${\bm{{\rm U}}}'$ and ${\bm{{\rm V}}}'$ as ${\bm{{\rm U}}}'' = {\bm{{\rm U}}}'{\bm{{\rm T}}}$ and ${\bm{{\rm V}}}'' = {\bm{{\rm V}}}'{\bm{{\rm T}}}$, respectively.

\partitle{Activation} Experiments show that sigmoid function $\sigma(\;)$ is the best choice for LCFN: ${\bm{{\rm U}}}'''\!=\!\sigma({\bm{{\rm U}}}'')$ and ${\bm{{\rm V}}}'''\!=\!\sigma({\bm{{\rm V}}}'')$.

LCFN contains $L$ layers and we use upper corner mark $(l)$ to indicate parameters for the $l$-th layer. The input of each layer is the output of the previous layer, thus we have:
\begin{small}
\begin{flalign}
\label{equ:convolution}
\left\{
\begin{array}{lcl}
{\bm{{\rm U}}}^{(l)} = \sigma\left(\bar{\bm{{\rm P}}}diag\left({{\bm{{\rm k}}}^U}^{(l)}\right)\bar{\bm{{\rm P}}}^\mathsf{T}{\bm{{\rm U}}}^{(l-1)}{\bm{{\rm T}}}^{(l)}\right) \\
{\bm{{\rm V}}}^{(l)} = \sigma\left(\bar{\bm{{\rm Q}}}diag\left({{\bm{{\rm k}}}^V}^{(l)}\right)\bar{\bm{{\rm Q}}}^\mathsf{T}{\bm{{\rm V}}}^{(l-1)}{\bm{{\rm T}}}^{(l)}\right)
\end{array}.
\right.
\end{flalign}
\end{small}After $L$ convolutional layers, we obtain embeddings in different level: $\left\{{\bm{{\rm U}}}^{(l)},{\bm{{\rm V}}}^{(l)}\right\}_{l=0\cdots L}$. We concatenate them as the \textit{predictive embeddings} to give the final prediction:
\begin{small}
	\begin{flalign}
	\label{equ:predict}
	\hat{\bm{{\rm R}}}=\left[{\bm{{\rm U}}}^{(0)},\cdots,{\bm{{\rm U}}}^{(L)}\right] \left[{\bm{{\rm V}}}^{(0)},\cdots,{\bm{{\rm V}}}^{(L)}\right]^\mathsf{T}.
	\end{flalign}
\end{small}To emphasize the effectiveness of our low-pass convolution, we choose the most simple yet widely-used way for embedding fusion (concatenation) and combination (inner product) \cite{NGCF,SCF,GNN_MatCom}. Other advanced fusion and combination functions \cite{NCF} are left to explore in future work.

\subsection{Optimization and Prediction}
To optimize our proposed model, we employ the BPR loss \cite{BPR}. We construct pairwise training set  $\mathcal{D}\!=\!\{(u,i,j)|{\bm{{\rm R}}}_{ui}\!=\!1 \bigwedge {\bm{{\rm R}}}_{uj}\!=\!0\}$ to learn the model. For a tuple $(u,i,j)\in \mathcal{D}$, we maximize the margin between $u$'s preference towards $i$ and $j$ by minimizing the BPR loss:
\begin{small}
	\begin{align}
	\mathcal{L} \!= \!\!\!\!\!\!\sum_{(u,i,j)\in\mathcal{D}}\!\!\!\!\!\!\!-\ln{\sigma\!\left(\hat{\bm{{\rm R}}}_{ui}\!-\!\hat{\bm{{\rm R}}}_{uj}\right)}\!+\!\frac{\lambda}{2}\!\left(\left\|{\bm{{\rm U}}}^{(0)}\right\|_F^2\!\!+\!\left\|{\bm{{\rm V}}}^{(0)}\right\|_F^2\!\!+\!\left\|{\bm{{\rm \Theta}}}\right\|_F^2\!\right), \nonumber
	\end{align}
\end{small}where $\hat{\bm{{\rm R}}}$ is given in Equation (\ref{equ:predict}). ${\bm{{\rm U}}}^{(0)}$, ${\bm{{\rm V}}}^{(0)}$ and ${\bm{{\rm \Theta}}} = \left\{{{\bm{{\rm k}}}^U}^{(l)}, {{\bm{{\rm k}}}^V}^{(l)}, {\bm{{\rm T}}}^{(l)}\right\}_{l=1, \cdots, L}$ are trainable parameters. We learn them by minimizing $\mathcal{L}$ with Adam \cite{adam}. When recommending for the user $u$, we rank items by $\hat{\bm{{\rm R}}}_u$ and return the top items to $u$. Of special notice is that by exploring the mean squared error (MSE) loss, our model can be extended to explicit feedbacks directly, which also suffer from the exposure and quantization noise.

\subsection{Complexity Analysis}
\label{subsec:complexity}
\textbf{Eigen-decomposition.} Original graph convolution requires all eigen-vectors of Laplacian matrices ${\bm{{\rm L}}}^U$ and ${\bm{{\rm L}}}^I$ as the transform bases, which costs $O(M^3+N^3)$ time. In our low-pass convolution, only front $\varPhi$/$\varPsi$ eigen-vectors are needed. For sparse matrices ${\bm{{\rm L}}}^U$ and ${\bm{{\rm L}}}^I$, the time complexity could be $O(M\varPhi^2+m\varPhi+N\varPsi^2+n\varPsi)$ with Lanczos method \cite{lanczos,lanczos1}, where $m$/$n$ is the number of nonzero elements of ${\bm{{\rm L}}}^U$/${\bm{{\rm L}}}^I$. We cannot omit the linear term $m\varPhi+n\varPsi$ here because hypergraph Laplacian matrices are not so sparse, and $m\varPhi/n\varPsi$ is larger than $M\varPhi^2/N\varPsi^2$ when $\varPhi/\varPsi$ is small.

\textbf{Model training.} Low-pass convolution also makes training more efficient. For the original graph convolution, ${\bm{{\rm U}}}' = {\bm{{\rm P}}}diag({\bm{{\rm k}}}^U){\bm{{\rm P}}}^\mathsf{T}{\bm{{\rm U}}}$  and ${\bm{{\rm V}}}' = {\bm{{\rm Q}}}diag({\bm{{\rm k}}}^V){\bm{{\rm Q}}}^\mathsf{T}{\bm{{\rm V}}}$ cost $O(K(M^2+N^2))$ time due to the multiplication with graph Fourier transform bases. In LCF, we use $\bar{\bm{{\rm P}}}$/$\bar{\bm{{\rm Q}}}$ to replace ${\bm{{\rm P}}}$/${\bm{{\rm Q}}}$, the time complexity reduces to $O(K(M\varPhi+N\varPsi))$.

\subsection{Comparison with Existing Models}
\label{subsec:comparison}
In this subsection we compare our model with some widely-explored GNNs \cite{GNN_WebScale, GNN_MatCom, NGCF} and GCNs \cite{SCF,GCN_rec1}.

\partitle{Comparison with GNNs} GNNs propagate signal (usually node embeddings) in the graph by the propagation matrix ${\bm{{\rm D}}}^{-1}{\bm{{\rm A}}}$ or ${\bm{{\rm D}}}^{-\frac{1}{2}}{\bm{{\rm A}}}{\bm{{\rm D}}}^{-\frac{1}{2}}$ (left or symmetric normalization, where ${\bm{{\rm D}}}$ denotes the degree of the nodes and ${\bm{{\rm A}}}$ is the adjacent matrix), and enhance the representation of each node with its neighbours. The propagation matrix contains one-hop connections thus the signal ${\bm{{\rm S}}}$ is propagated to one-hop neighbours by ${\bm{{\rm D}}}^{-1}{\bm{{\rm A}}}{\bm{{\rm S}}}$ in one GNN layer. Propagated by $L$ layers, the signal of a certain node is enhanced by signals of its neighbours within $L$ hops. 

In each layer of our LCFN model, the convolution of the signal ${\bm{{\rm S}}}$ is $\bar{\bm{{\rm P}}}diag({\bm{{\rm k}}})\bar{\bm{{\rm P}}}^\mathsf{T}{\bm{{\rm S}}}$, where $\bar{\bm{{\rm P}}}$ is the first $\varPhi$ eigen-vectors of the Laplacian matrix and widely used as the node embedding \cite{spec_clus}. If we fix ${\bm{{\rm k}}}$ to all-one vector ${\bm 1}$ (i.e., we only filter the signal by LCF without convolution), $\bar{\bm{{\rm P}}}\bar{\bm{{\rm P}}}^\mathsf{T}$ is an advanced propagation matrix, where the propagations are weighted based on the connection strength by taking all-hop connections into consideration \cite{SPLR}. Even without graph convolution, LCFN still works better than GNN. By learning the convolutional kernel, our LCFN gains further enhancement.

\partitle{Comparison with GCNs} All existing GCNs avoid eigen-decomposition by simplifying graph convolution. \citet{GCN} first proposed an efficient GCN with the Chebyshev polynomial and \citet{GCN_rec2} leveraged this model in recommendation tasks. The convolutional kernel is reconstructed by a $K$-order polynomial: ${\bm{{\rm k}}} = \sum_{k=0}^{K}\theta_kT_k({\bm{{\rm \Lambda}}})$ and the convolution of the signal ${\bm{{\rm S}}}$ is ${\bm{{\rm P}}}\Big(\sum_{k=0}^{K}\theta_kT_k({\bm{{\rm \Lambda}}})\Big) {\bm{{\rm P}}}^\mathsf{T}{\bm{{\rm S}}} = \sum_{k=0}^{K}\theta_kT_k({\bm{{\rm L}}}){\bm{{\rm S}}}$. In this way, we do not need to construct ${\bm{{\rm P}}}$ and ${\bm{{\rm \Lambda}}}$. Since the Laplacian matrix ${\bm{{\rm L}}} = {\bm{{\rm I}}}-{\bm{{\rm D}}}^{-\frac{1}{2}}{\bm{{\rm A}}}{\bm{{\rm D}}}^{-\frac{1}{2}}$ and $T_k({\bm{{\rm L}}})$ is a $k$-order polynomial of ${\bm{{\rm L}}}$, the convolution becomes $p_K({\bm{{\rm D}}}^{-\frac{1}{2}}{\bm{{\rm A}}}{\bm{{\rm D}}}^{-\frac{1}{2}}){\bm{{\rm S}}}$, where $p_K(x)$ is the $K$-order polynomial of $x$. In this case, the simplified graph convolution is a linear combination of propagations within $K$ hops. We can see that this GCN is indeed an advanced version of GNN, which can be achieved better by LCFN with only LCF. Another issue is that the density of $T_k({\bm{{\rm L}}})$ increases exponentially with the increasing of $k$, thus the order is very limited, or the convolution will be very space- and time-consuming.

A further simplification is proposed by \citet{semi_super} and \citet{GCN_rec2,SCF} then leveraged it in recommendation. \citet{semi_super} fixed the kernel as $\theta(2{\bm{{\rm I}}}-{\bm{{\rm \Lambda}}})$, thus the convolution of ${\bm{{\rm S}}}$ is ${\bm{{\rm P}}}\theta(2{\bm{{\rm I}}} - {\bm{{\rm \Lambda}}}) {\bm{{\rm P}}}^\mathsf{T}{\bm{{\rm S}}} = ({\bm{{\rm I}}} + {\bm{{\rm D}}}^{-\frac{1}{2}} {\bm{{\rm A}}} {\bm{{\rm D}}}^{-\frac{1}{2}}){\bm{{\rm S}}}\theta$. It is obvious that GCN degenerates to GNN thoroughly: The convolution of the signal is the combination of the signal itself and signals from one-hop neighbours. The parameter $\theta$ is regarded as the kernel and extended to a matrix ${\bm{{\rm \Theta}}}$, nevertheless it is actually a transformation matrix widely used in GNNs \cite{GNN_MatCom, GNN_WebScale, NGCF} as well as in our model (matrix ${\bm{{\rm T}}}$). Some GNNs \cite{GNN_GCN_comp1,GNN_GCN_comp2} are declared equivalent to GCNs, unfortunately, the truth is by following \citet{semi_super}, most GCNs degenerate to GNNs.

In this paper, we provide a new strategy to simplify the computation. We only need to construct front eigen-vectors of the Laplacian matrix, which can be solved in acceptable time. In existing work, the kernel is not trainable \cite{semi_super,GCN_rec2,SCF} or partially trainable \cite{GCN,GCN_rec2}, which cripples the graph convolution seriously. In our LCFN model, the convolutional kernel is fully trained from the data. Another point to support that our model is more powerful is that: in all existing GCNs, the signal ${\bm{{\rm S}}}$ is multiplied with a sparse matrix, while in LCFN, the signal ${\bm{{\rm S}}}$ is multiplied with a dense convolution operator.

\section{Experiment}
\label{sec:experiment}
Experiments are conducted in this section to demonstrate the effectiveness and efficiency of LCFN by comparing it against several state-of-the-art models on two real-world datasets. We focus on answering three research questions:

\partitle{RQ1} How is the performance of our LCFN model?

\partitle{RQ2} How is the efficiency improvement by LCF?

\partitle{RQ3} How is the effectiveness improvement by LCF?

\subsection{Experimental Setup}
\label{subsec:exp_setup}

\partitle{Datasets} We adopt two real-world datasets, \textit{Amazon} and \textit{Movielens}, to evaluate our model.

\begin{table}[ht]
    \caption{Statistics of datasets}
    \centering
    \label{tab:datasets}
    \scalebox{1}{
    \begin{tabular}{ccccc}
        \toprule[1.2pt]
                Dataset & Purchase & User & Item & Sparsity\\
        \hline
        \textit{Amazon} & 347,393 & 20,247 & 11,589 & 99.8519\% \\
        \textit{Movielens} & 995,154 & 6,022 & 3,043 & 94.5694\% \\
        \bottomrule[1.2pt]
    \end{tabular}}\vspace{-2mm}
\end{table}

\vspace{-2mm}
\begin{itemize}
	\item \textbf{Amazon.} This \textit{Amazon} dataset \cite{Amazon} is the user reviews collected from the E-commerce website \textit{Amazon.com}. In this paper we adopt the \textit{Electronic} category.
    \vspace{-1mm}
    \item \textbf{MovieLens.} This \textit{Movielens} dataset \cite{Movielens} is collected through the movie website \textit{movielens.umn.edu} and we use the 1M version.
\end{itemize}
\vspace{-2mm}

These two datasets are all explicit feedbacks, we set the interaction $(u,i)$ as ``1'' if $u$ rated $i$ and ``0'' otherwise. We filter \textit{Amazon} and \textit{Movielens} with 10- and 20-core respectively (when filtered with $n$-core, we remove users and items within $n$ interactions) to reduce the sparsity. Datasets after filtering are shown in Table \ref{tab:datasets}. 
Each dataset is split into training set (80\%), validation set (10\%), and test set (10\%) randomly. 
We train models on the training set, determine all hyperparameters on the validation set, and report the performances on the test set.

\partitle{Baselines} We adopt the following methods as baselines for performance comparison.
\vspace{-2mm}

\begin{itemize}
\item{\textbf{MF:} This \textbf{M}atrix \textbf{F}actorization method encodes users and items with embeddings by factorizing the observed matrix \cite{MF}. It is a basic yet competitive baseline for item recommendation.}
\vspace{-1mm}

\item{\textbf{GCMC:} This \textbf{G}raph \textbf{C}onvolutional \textbf{M}atrix \textbf{C}omple-tion is a GNN-based recommendation model which utilizes a graph auto-encoder to propagate features to neighbours \cite{GNN_MatCom}. In our experiments, features are learnable embeddings and we set $R=1$ for implicit feedbacks.}
\vspace{-1mm}

\item{\textbf{NGCF:} This \textbf{N}eural \textbf{G}raph \textbf{C}ollaborative \textbf{F}iltering method is a GNN model which propagates embeddings through the graph to refine the representation of users and items \cite{NGCF}.}
\vspace{-1mm}

\item{\textbf{SCF:} This \textbf{S}pectral \textbf{C}ollaborative \textbf{F}iltering method is the state-of-the-art GCN model for recommendation \cite{SCF}. The convolutional kernel is fixed to $\theta({\bm{{\rm I}}}+{\bm{{\rm \Lambda}}})$. By extending $\theta$ to ${\bm{{\rm \Theta}}}$, the convolution of a signal ${\bm{{\rm S}}}$ is $(2{\bm{{\rm I}}}-{\bm{{\rm D}}}^{-1}{\bm{{\rm A}}}){\bm{{\rm S}}}{\bm{{\rm \Theta}}}$.}
\vspace{-1mm}

\item{\textbf{CGMC:} This \textbf{C}onvolutional \textbf{G}eometric \textbf{M}atrix \textbf{C}ompletion model is also a GCN model, which leverages graph convolution to extract features \cite{GCN_rec1}. The kernel is set to $\theta({\bm{{\rm I}}}-(1-\sigma){\bm{{\rm \Lambda}}})$ and the convolution of a signal ${\bm{{\rm S}}}$ is $(\sigma{\bm{{\rm I}}}+(1-\sigma){\bm{{\rm D}}}^{-\frac{1}{2}} {\bm{{\rm A}}} {\bm{{\rm D}}}^{-\frac{1}{2}}){\bm{{\rm S}}}{\bm{{\rm \Theta}}}$, where $\sigma$ is a hyperparameter to balance self signal and neighbour signals.}
\end{itemize}
\vspace{-2mm}

Noting that MF, GCMC, CGMC are designed for explicit feedbacks, we use BPR \cite{BPR} to optimize all models in our experiments.

\partitle{Evaluation Protocols} To evaluate the performances of our proposed model and baselines, we rank all items for each user in validation/test set and recommend the top-2 items to the user. We then adopt two metrics, $F_1$-score@2 and normalized discounted cumulative gain (NDCG@2) to evaluate the performance (results for top-$\{2,5,10,20,50,100\}$ recommendation can be found in Appendix \ref{app:experiment_results}). $F_1$-score is extensively used to test the accuracy of binary classifier and NDCG is a position-sensitive metric widely used to measure the ranking quality. We recommend top-2 items to each user and calculate metrics, and finally use the average metrics of all users as the model performance.

\partitle{Parameter Setting} To compare fairly, all models in our experiments are tuned with the following strategies: The maximum iteration number is set to 200. In each iteration, we enumerate all positive samples to train the model and then test it. The learning rate $\eta$ and the regularization coefficient $\lambda$ are determined by grid searching in the coarse grain range of $\{0.0001,0.001,0.01\}\otimes \{0.001,0.01, 0.1\}$ and then in the fine grain range, which is based on the result of coarse tuning. For example, if a certain model achieves the best performance when $\eta=0.01$ and $\lambda=0.1$, we then tune it in the range of $\{0.002, 0.005, 0.01, 0.02, 0.05\} \otimes \{0.02, 0.05, 0.1, 0.2, 0.5\}$. The batch size is set as 10,000 to utilize GPU. We evaluate different embedding dimensionality $D$ in the range of $\{8,16,\cdots,128\}$. The layer number $L$ is determined in the range of $\{1,2,\cdots,5\}$. To explore the best passband cut-off frequency, we define $F=\frac{\varPhi}{M}=\frac{\varPsi}{N}$, and tune our model with respect to $F$ in the range of $\{0.0002,0.0005,0.001,\cdots,0.2\}$. The experiments are repeated 5 times for each parameter setting in model tuning and 10 times for testing. The embedding layer of all deep models in our experiments are initialized by pretrained MF.

\subsection{Performance of LCFN (RQ1)}

Performances of all models on two datasets are show in Table \ref{tab:exper_result}. ``Imp'' means relative improvement and ``BB'' means the best baseline. In GNNs and GCNs, the graph is used to provide connection information, thus we should gain performance improvement over MF. However, to our surprise, MF is a very competitive baseline and GNN/GCN baselines only outperform it very slightly when compared fairly. In these baselines, GCMC and NGCF are GNNs while SCF and CGMC are GCNs. As we discussed, all existing GCNs lose the ability of graph convolution and degenerate to GNNs, or advanced version of GNN but with the same mechanism. This viewpoint is also supported by the experiments --- SCF and CGMC perform almost the same with GCMC and NGCF. Utilizing unscathed graph convolution to extract features, LCFN gains significant improvement over existing GNNs and GCNs: LCFN outperforms BB 16.18\% on $F_1$-score and 17.71\% on NDCG for the best case.
\vspace{-2mm}
\begin{table}[ht]
	\caption{Recommendation performance (test set)}
	\centering
	\label{tab:exper_result}
	\scalebox{0.9}{
		\begin{tabular}{cc|cc|cc}
			\hline
			\multicolumn{2}{c|}{\multirow{2}{*}{Model}} & \multicolumn{2}{c|}{\textit{Amazon}} & \multicolumn{2}{c}{\textit{Movielens}} \\
			\cline{3-6}
			& & $F_1$-score & NDCG & $F_1$-score & NDCG \\
			\hline	 	 
			\multicolumn{2}{c|}{MF} & 0.01356 & 0.01742 & 0.07506 & 0.25100\\
			\multicolumn{2}{c|}{GCMC} & 0.01377 & 0.01742 & 0.07689 & \textbf{0.26316} \\
			\multicolumn{2}{c|}{NGCF} & 0.01349 & 0.01733 & 0.07453 & 0.25436 \\
			\multicolumn{2}{c|}{SCF} & 0.01424 & 0.01787 & 0.07808 & 0.26173 \\
			\multicolumn{2}{c|}{CGMC} & 0.01363 & 0.01728 & 0.07705 & 0.25790 \\
			\multicolumn{2}{c|}{LCFN} & \textbf{0.01654} & \textbf{0.02104} & \textbf{0.08151} & 0.26129 \\
			\hline
			\multirow{2}{*}{Imp} & MF & 22.00\% & 20.77\% & 8.60\% & 4.10\% \\
			& BB & 16.18\% & 17.71\% & 4.40\% & -0.71\% \\
			\hline
	\end{tabular}}\vspace{-2mm}
\end{table}

Comparing CGMC and SCF, we can see that the kernel $\theta({\bm{{\rm I}}}+{\bm{{\rm \Lambda}}})$ is better than $\theta({\bm{{\rm I}}}-(1-\sigma){\bm{{\rm \Lambda}}})$. For existing GCNs, since the convolutional kernels are fixed, we need to design better kernels based on our experience yet the improvement is usually very marginal. In the computer vision domain, the evolution from hand-crafted features such as Histogram of Oriented Gradient (HOG) \cite{HOG} to end-to-end models such as CNN \cite{alexnet} shows that the power of CNN is derived from learning the kernels from the data. In our LCFN, we leverage the original graph convolution and play its roles by designing an end-to-end GCN. Experiments also show that our learnable kernels are more powerful than the hand-crafted kernels.


\subsection{Model Tuning}
\begin{figure*}[t]
	\centering
	\subfigure[Impact of $D$]{
		\includegraphics[scale = 0.2]{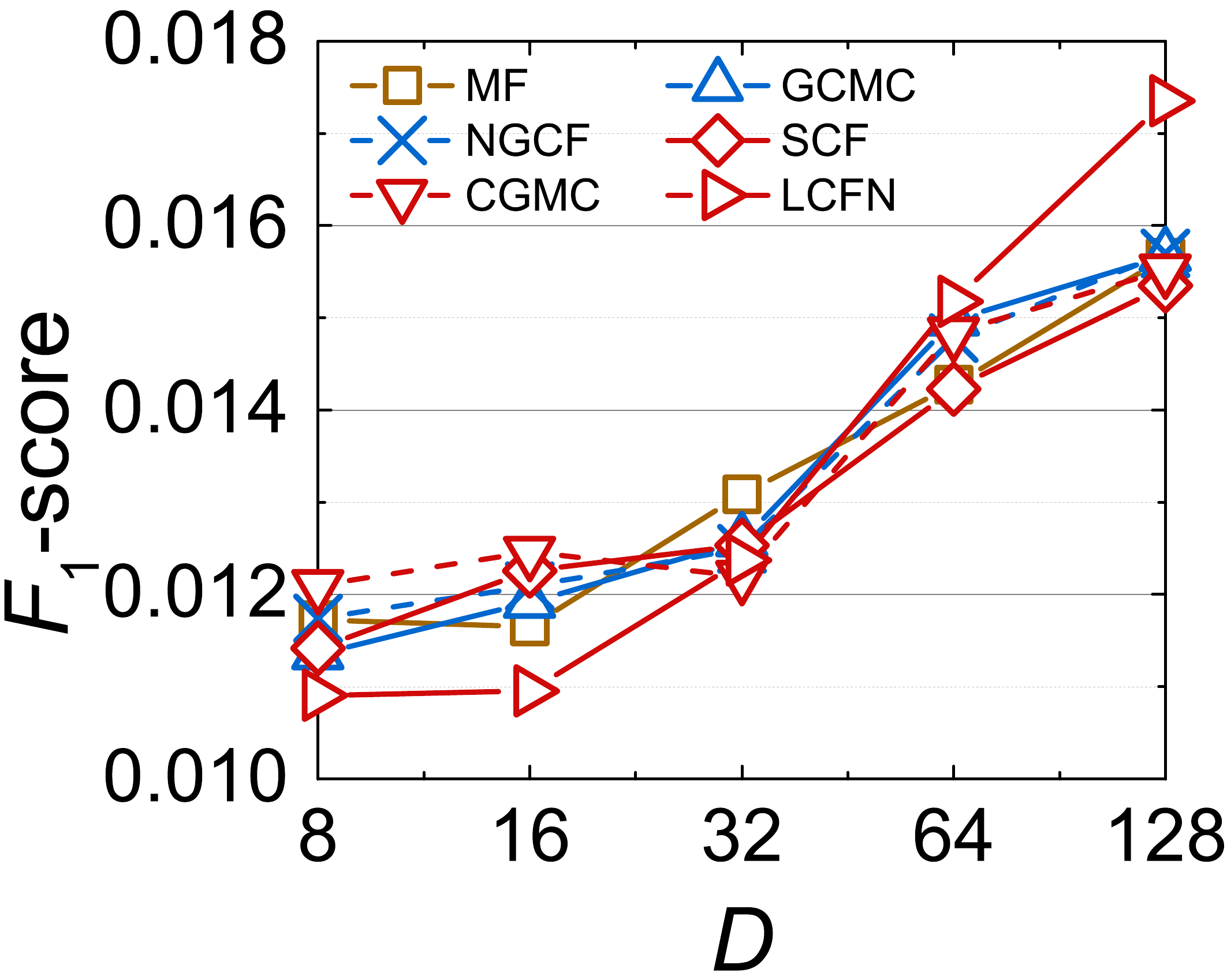}
		\label{subfig:K}
	}
	\hspace{5mm}
	\subfigure[Impact of $L$]{
		\includegraphics[scale = 0.2]{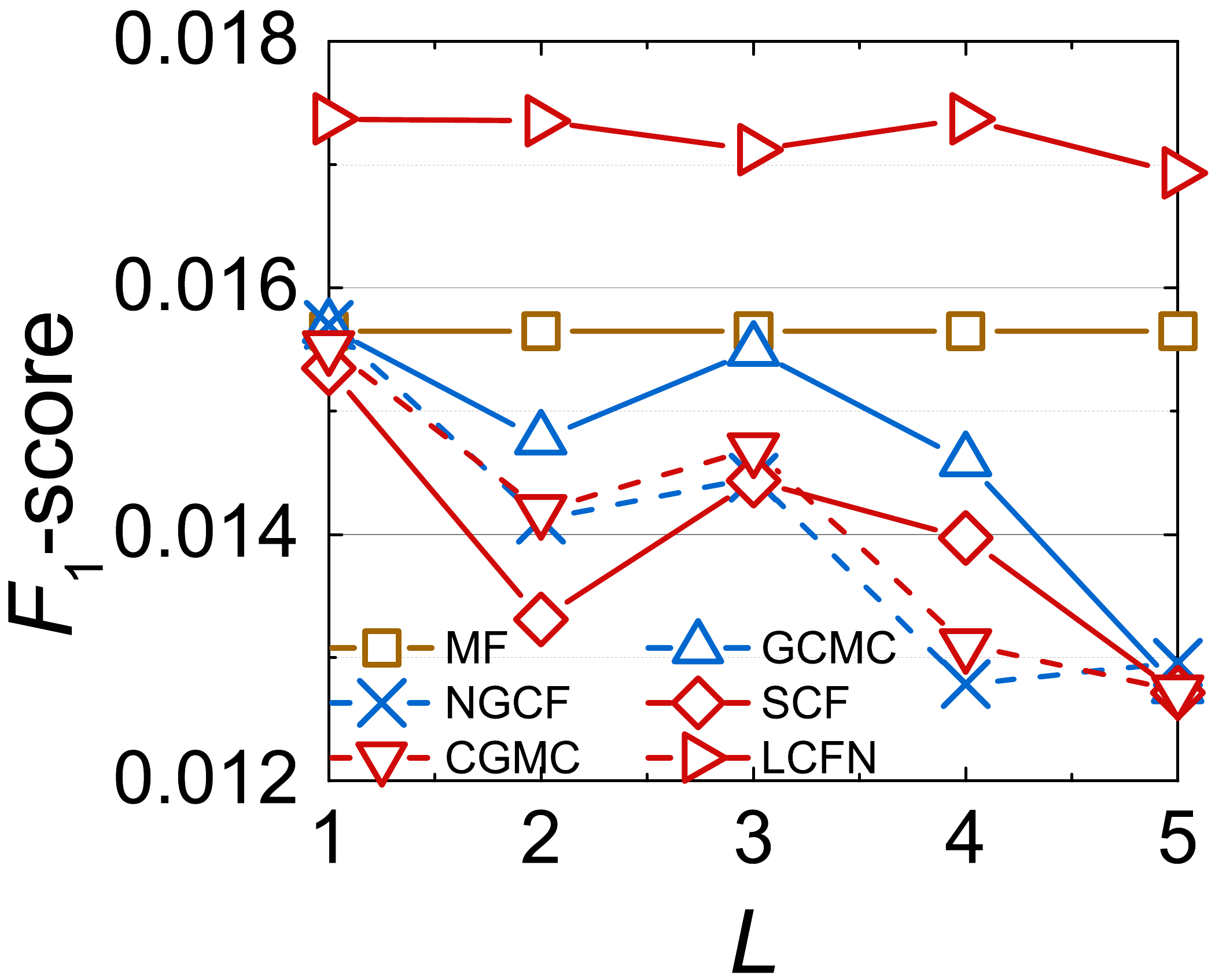}
		\label{subfig:L}
	}
	\hspace{5mm}
	\subfigure[Impact of $F$]{
		\includegraphics[scale = 0.2]{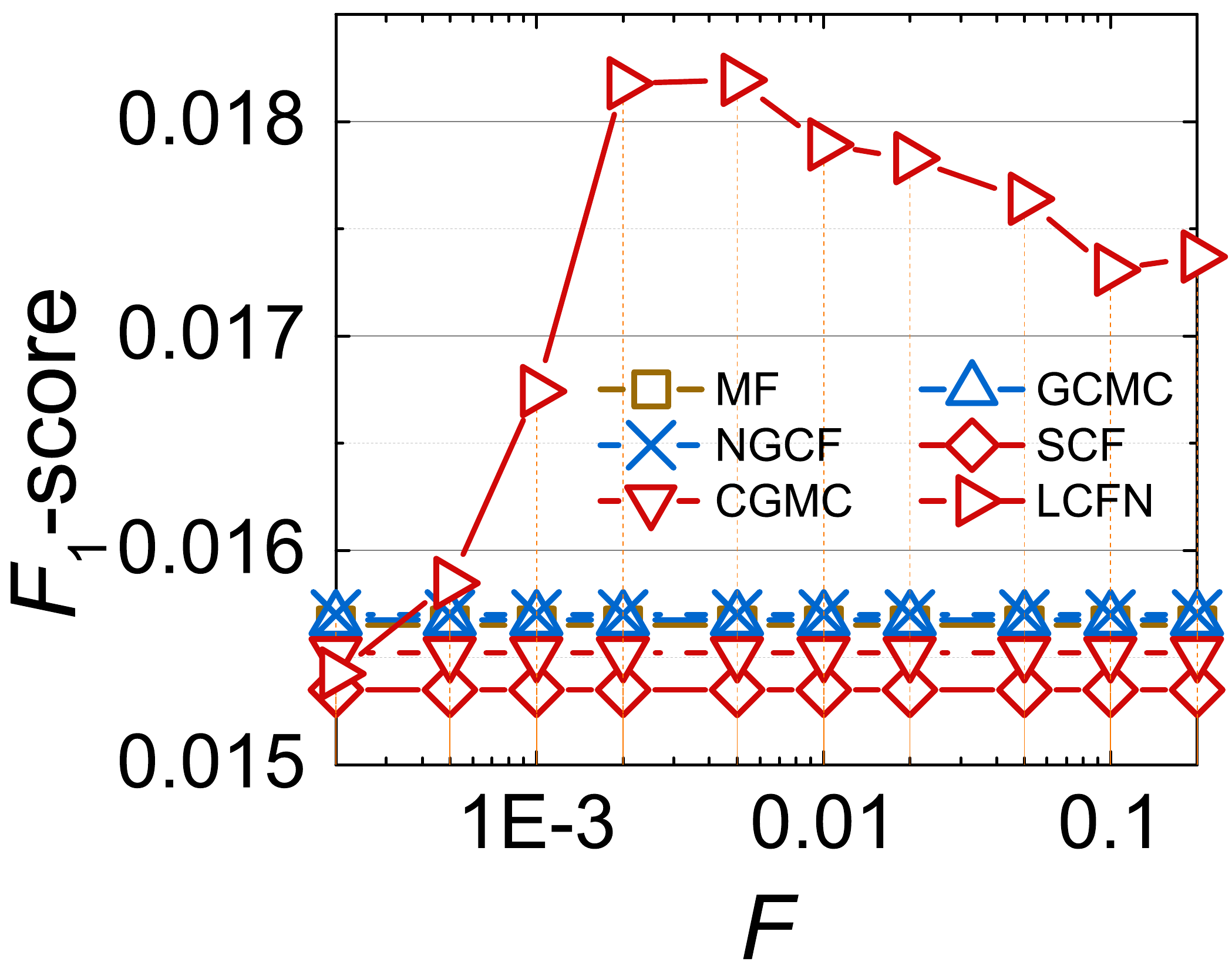}
		\label{subfig:F}
	}
	\caption{Model tuning (\textit{Amazon}, validation set).}\vspace{-2mm}
	\label{fig:model_tuning}
\end{figure*}

The result of model tuning is shown in Figure \ref{fig:model_tuning}. For a fair comparison, we tune all models with the same strategy introduced in Subsection \ref{subsec:exp_setup}. Due to the limited space, we only report experiments on \textit{Amazon} dataset on account of the similar performances on the two datasets. To make sure all models achieve their best performance for each hyperparameter setting, we retune them with respect to $\eta$ and $\lambda$ and initialize the deep models with pretrained MF for each $D$, $L$, and $F$ in Figure \ref{fig:model_tuning}.


Figure \ref{subfig:K} illustrates how the predictive embedding dimensionality $D$ impacts the models. Denoting $K$ is the embedding dimensionality of the input layer and $L$ is the depth of the models, we define $D=K$ for MF and $D=(L+1)K$ for all GNNs/GCNs, where the predictive embeddings are the concatenation of embeddings from the input layer and $L$ convolutional/propagation layers. As shown in the figure, we gain better performance by increasing the representation ability. In this tuning procedure, we fix $L=1$ and $F=0.2$.

The number of layers $L$ is analyzed in Figure \ref{subfig:L}. \citet{GNN_MatCom} fixed $L$ to 1 in GCMC yet it is an experimental conclusion depending on specific datasets, thus we also explore the best $L$ for GCMC in our experiments. In GNNs and GCNs, with the increasing of $L$, $D$ also increases, thus we are not sure if the improvement is caused by the increasing of $L$ or of $D$. To control variables, we fix $D$ to 128 when setting different $L$. From Figure \ref{subfig:L} we can see that the properties of GNN and GCN baselines are pretty similar, this is also experimental evidence of our key viewpoint: existing GCNs degenerate to GNNs. LCFN shows different properties because we really perform convolution rather than propagating embeddings. Figure \ref{subfig:L} also shows that high-order propagation is harmful, which is supported by \citet{GNN_MatCom}. \citet{NGCF} got the different conclusion because they did not control variables. As LCFN gains no improvement with a deeper structure, we set $L=1$ for efficiency. In this procedure, we fix $D=128$ and $F=0.2$.

Figure \ref{subfig:F} shows the sensitivity analysis of the cut-off frequencies $F$. When $F$ is too small, the signal is over-filtered and much useful information is corrupted. However, when $F$ is too large, the noise is reserved hence the performance also decreases. LCFN achieves the best performance when $F = 0.005$. In this procedure, we fix $D=128$ and $L=1$.

\begin{table}[ht]
    \caption{Effectiveness of pretraining ($F_1$-score, validation set)}
    \centering
    \label{tab:pretrain}
    \scalebox{0.755}{
    \begin{tabular}{c|ccc|ccc}
        \hline
        \multirow{2}{*}{Model} & \multicolumn{3}{c|}{\textit{Amazon}} & \multicolumn{3}{c}{\textit{Movielens}} \\
        \cline{2-7}
        & No-pre & Pre & Imp & No-pre & Pre & Imp \\
        \hline
        GCMC & 0.01466 & 0.01568 & 6.91\% & 0.07288 & 0.07828 & 7.42\% \\
        NGCF & 0.01190 & 0.01570 & 31.93\% & 0.07031 & 0.07877 & 12.03\% \\
        SCF & 0.01149 & 0.01535 & 33.59\% & 0.07300 & 0.07760 & 6.29\% \\
        CGMC & 0.01265 & 0.01552 & 22.69\% & 0.07169 & 0.07700 & 7.41\% \\
        LCFN & 0.01743 & 0.01819 & 4.38\% & 0.08013 & 0.07994 & -0.24\% \\
        \hline
    \end{tabular}}\vspace{-4mm}
\end{table}

In our experiments, all deep models are pretrained to get better performance. Here we also discuss the effectiveness of pretraining. In Table \ref{tab:pretrain}, ``No-pre'' means no pretraining, ``Pre'' means pretraining, and ``Imp'' means relative improvement. Please note that we also retune all models to make sure they achieve their best performances without pretraining. Compared with baselines, another advantage of LCFN is that it dose not rely on pretraining. 

Comparing with Tables \ref{tab:exper_result} and \ref{tab:pretrain}, we can observe an interesting phenomenon. NDCG performs pretty well on the validation set yet performs poorly on the test set, while for SCF, we face the opposite situation. Compared with the basic GNN model GCMC, NGCF additionally propagates signals based on element-wise preference similarity, hence shows better performance on the validation set. However, this strategy may make the model easily get overfitted to the validation set. Benefiting from the well-designed kernel, SCF shows better generalization and performs the best among all baselines on the test set.

\subsection{Efficiency Enhancement by LCF (RQ2)}
\label{subsec:time}
As we analyzed in Subsection \ref{subsec:complexity}, LCF improves the efficiency of GCN. In this subsection, we report some experimental evidence.

\begin{figure*}[ht]
	\centering
	\subfigure[Time consumption for eigen-decomposition]{
		\includegraphics[scale = 0.2]{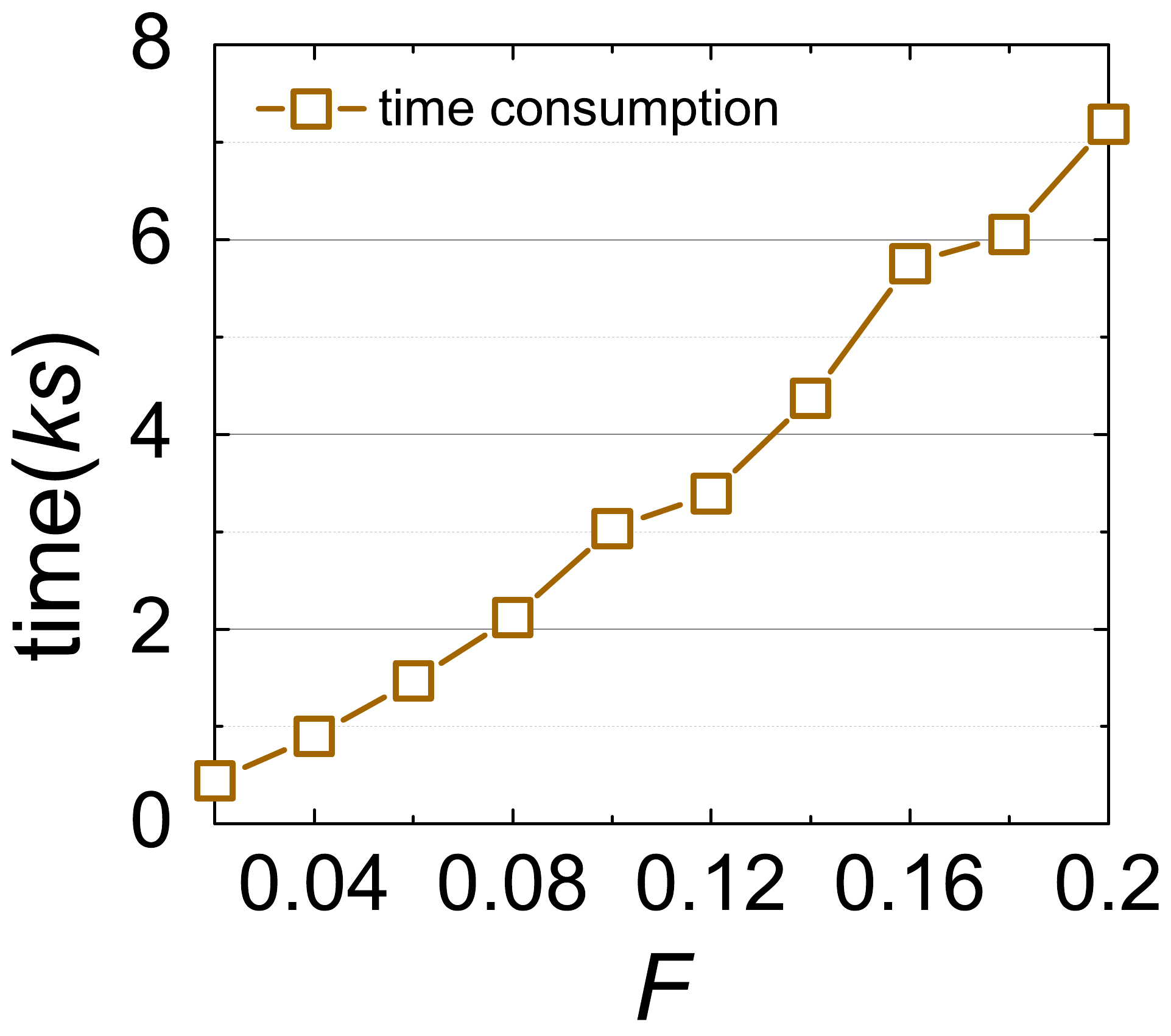}
		\label{subfig:decom_time}
	}
	\hspace{5mm}
	\subfigure[Time consumption for model training]{
		\includegraphics[scale = 0.2]{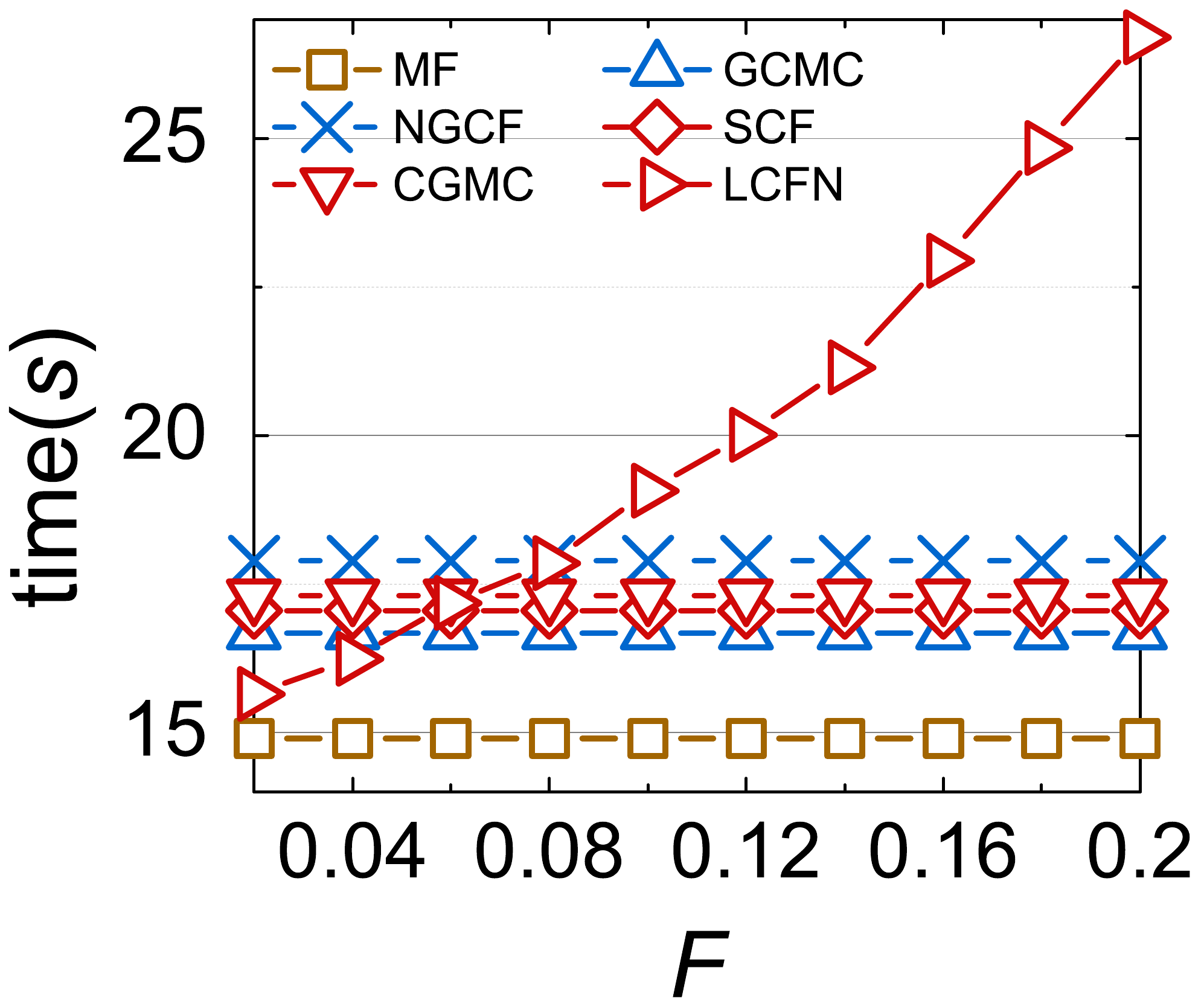}
		\label{subfig:train_time}
	}
	\hspace{5mm}
	\subfigure[Models in the time-accuracy space (validation set)]{
		\includegraphics[scale = 0.2]{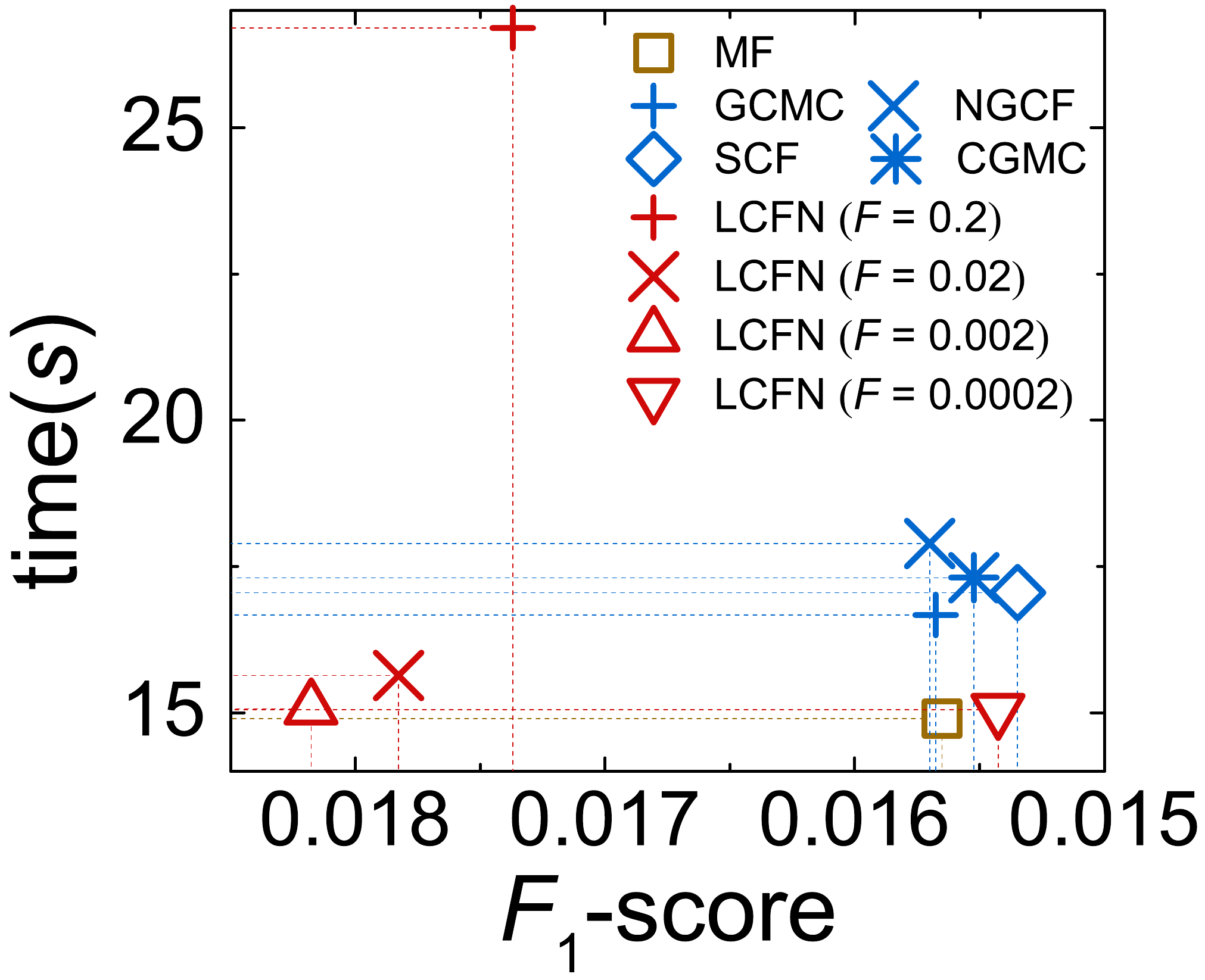}
		\label{subfig:skyline}
	}
	\caption{Efficiency and Effectiveness improvement of LCF.}\vspace{-2mm}
	\label{fig:time}
\end{figure*}

The efficiency improvement are caused by two aspects: constructing transform bases (eigen-decomposition), and model training. Figure \ref{subfig:decom_time} shows the time consumption of eigen-decomposition with different $F$. The time complexity is $O((M^3+N^3)F^2+(mM+nN)F)$, and Figure \ref{subfig:decom_time} shows that the time complexity is mainly determined by the linear term of $F$. As shown in Figure \ref{subfig:F}, LCFN achieves the best performance when $F=0.005$. In this case, we can save considerable time. The time consumption of one iteration during training is illustrated in Figure \ref{subfig:train_time}. The training efficiency of LCFN is also satisfactory: When $F=0.005$, LCFN is obviously faster than all GNN/GCN baselines.

Since the datasets are too large, the original graph convolution ($F=1$) is inapplicable. However, from the trend of the curves in Figures \ref{fig:time}(a)(b), we can infer that we improve the efficiency significantly than the GCN with the original graph convolution.

\subsection{Effectiveness Enhancement by LCF (RQ3)}

The effectiveness improvement of LCF has been represented in Figure \ref{subfig:F}: We gain dramatic improvement by setting a small passband cut-off frequency. We further represent all models in the time-accuracy space in Figure \ref{subfig:skyline}. The horizontal axis represents the time consumption of one iteration in training, and the vertical axis represents $F_1$-score. Noting that $F_1$-score is in descending order, models in the lower left area are better. By tuning appropriate $F$, LCFN outperforms GNN/GCN baselines significantly in both time and accuracy aspects.

\section{Conclusion and Future Work}
\label{sec:conclusion}
In this paper, we first designed a LCF to remove the high-frequency exposure noise and quantization noise from the observed interaction matrix to uncover the underlying rating matrix. We then proposed the 2D graph convolution by extending convolution from the Euclidean domain to the graph domain. Finally, we designed a deep GCN and injected the LCF into it. LCF removes the noise to improve the prediction accuracy as well as provides a new way to design applicable graph convolution in an unscathed way.

For future work, we have interests in validating the effectiveness of LCFN in other tasks, such as social network, knowledge graph, and even computer vision, since images are also 2D low-frequency signals contaminated by high-frequency noise. Moreover, we want to inject side information into LCFN, such as user reviews, visual features. We also want to propose multi-graph LCFN. We construct graphs by such as tags \cite{tag} or item transfer Morkov chain \cite{markov}, and use these graphs for convolution in multiple channels.

\newpage

\bibliographystyle{icml2020}

\clearpage

\appendix

\label{app:lcfn_model}
\begin{figure*}[ht]
	\centering
	\includegraphics[scale = 0.5]{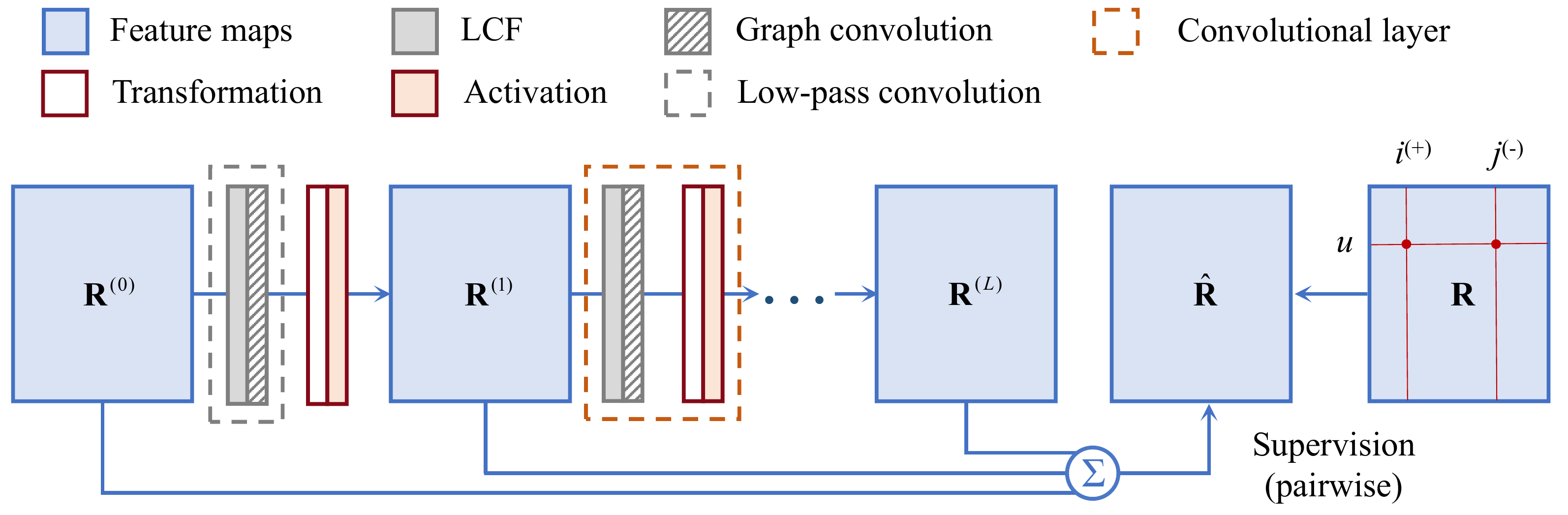}
	\caption{Illustration of LCFN model.}
	\label{fig:GCN}
\end{figure*}

\begin{table*}[ht]
	\caption{Recommendation performance (test set)}  
	\begin{center}  
		\label{tab:exper_result_all}
		\scalebox{1}{
			\begin{tabular}{m{1.3cm}<{\centering}||m{1.2cm}<{\centering}|c|cccccccc}  
				\hline
				\hline
				\multirow{2}{*}{Datasets} & \multicolumn{2}{c|}{\multirow{2}{*}{Metrics}} & \multirow{2}{*}{MF} & \multirow{2}{*}{GCMC} & \multirow{2}{*}{NGCF} & \multirow{2}{*}{SCF} & \multirow{2}{*}{CGMC} & \multirow{2}{*}{LCFN} & \multicolumn{2}{c}{Improvement} \\
				& \multicolumn{2}{c|}{} & & & & & & & MF & BB\\
				\hline
				\hline
				
				\multirow{10}{*}{\textit{Amazon}} & \multirow{5}{*}{$F$-1@} 
				& 2 & 0.01356 & 0.01377 & 0.01349 & 0.01424 & 0.01363 & \textbf{0.01654} & 22.00\% & 16.18\% \\
				& & 5 & 0.01484 & 0.01452 & 0.01442 & 0.01459 & 0.01472 & \textbf{0.01643} & 10.69\% & 10.69\% \\
				& & 10 & 0.01355 & 0.01318 & 0.01352 & 0.01366 & 0.01350 & \textbf{0.01465} & 8.09\% & 7.20\% \\
				& & 20 & 0.01155 & 0.01114 & 0.01128 & 0.01126 & 0.01112 & \textbf{0.01247} & 7.92\% & 7.92\% \\
				& & 50 & 0.00833 & 0.00817 & 0.00810 & 0.00803 & 0.00805 & \textbf{0.00902} & 8.34\% & 8.34\% \\
				& & 100 & 0.00619 & 0.00607 & 0.00609 & 0.00599 & 0.00604 & \textbf{0.00654} & 5.70\% & 5.70\% \\
				
				\cline{2-11}    
				& \multirow{5}{*}{NDCG@} 
				& 2 &0.01742 & 0.01742 & 0.01733 & 0.01787 & 0.01728 & \textbf{0.02104} & 20.77\% & 17.71\% \\
				& & 5 & 0.02354 & 0.02362 & 0.02294 & 0.02343 & 0.02323 & \textbf{0.02711} & 15.15\% & 14.79\% \\
				& & 10 & 0.02940 & 0.02945 & 0.02943 & 0.02983 & 0.02944 & \textbf{0.03373} & 14.70\% & 13.05\% \\
				& & 20 & 0.03686 & 0.03657 & 0.03628 & 0.03673 & 0.03626 & \textbf{0.04180} & 13.42\% & 13.42\% \\
				& & 50 & 0.04828 & 0.04820 & 0.04724 & 0.04776 & 0.04719 & \textbf{0.05445} & 12.77\% & 12.77\% \\
				& & 100 & 0.05865 & 0.05850 & 0.05808 & 0.05806 & 0.05760 & \textbf{0.06527} & 11.30\% & 11.30\% \\
				
				\hline
				\hline
				\multirow{10}{*}{\textit{Movielens}} & \multirow{5}{*}{$F$-1@} 
				& 2 & 0.07506 & 0.07689 & 0.07453 & 0.07808 & 0.07705 & \textbf{0.08151} & 8.60\% & 4.40\% \\
				& & 5 & 0.11531 & 0.11659 & 0.11533 & 0.11888 & 0.11789 & \textbf{0.12133} & 5.22\% & 2.06\% \\
				& & 10 & 0.14217 & 0.14372 & 0.14254 & 0.14513 & 0.14311 & \textbf{0.14822} & 4.26\% & 2.13\% \\
				& & 20 & 0.15728 & 0.15643 & 0.15815 & 0.16002 & 0.15727 & \textbf{0.16250} & 3.31\% & 1.55\% \\
				& & 50 & 0.15115 & 0.15028 & 0.15270 & 0.15262 & 0.14973 & \textbf{0.15576} & 3.05\% & 2.00\% \\
				& & 100 & 0.12843 & 0.12760 & 0.12976 & 0.12975 & 0.12671 & \textbf{0.13151} & 2.40\% & 1.34\% \\
				
				\cline{2-11}
				& \multirow{5}{*}{NDCG@} 
				& 2 & 0.25100 & \textbf{0.26316} & 0.25436 & 0.26173 & 0.25790 & 0.26129 & 4.10\% & -0.71\% \\
				& & 5 & 0.23368 & 0.24114 & 0.23666 & 0.24191 & 0.24083 & \textbf{0.24268} & 3.85\% & 0.32\% \\
				& & 10 & 0.23292 & 0.23614 & 0.23473 & 0.23982 & 0.23724 & \textbf{0.24285} & 4.26\% & 1.26\% \\
				& & 20 & 0.25052 & 0.24955 & 0.25111 & 0.25595 & 0.25140 & \textbf{0.26025} & 3.88\% & 1.68\% \\
				& & 50 & 0.30066 & 0.29718 & 0.30041 & 0.30406 & 0.29815 & \textbf{0.31237} & 3.89\% & 2.73\% \\
				& & 100 & 0.35339 & 0.34841 & 0.35279 & 0.35600 & 0.34906 & \textbf{0.36388} & 2.97\% & 2.21\% \\
				
				\hline
				\hline
		\end{tabular}} 
	\end{center} 
\end{table*}  

\section{An Example to Understand LCF in the Recommendation Context}
\label{app:example_GFT}
\begin{figure}[ht]
	\centering
	\includegraphics[scale = 0.6]{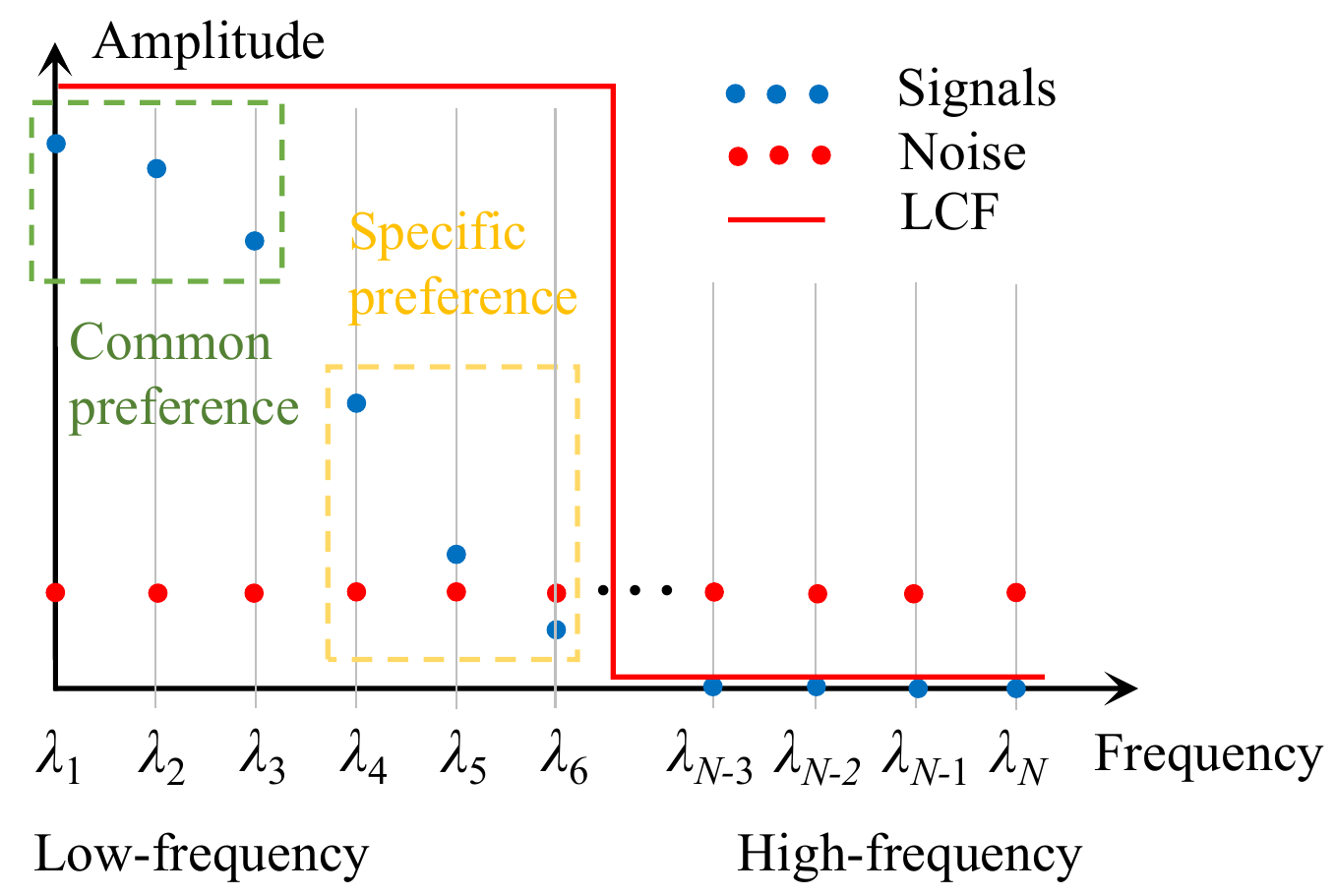}
	\caption{An intuitive examples for the mechanism of low-pass collaborative filter in recommendation.}
	\label{fig:intuitive_example}
\end{figure}

Here, we give an example about Low-pass Collaborative Filter (LCF) to understand it intuitively in the recommendation context. For concise illustration, we introduce a 1D graph data on the user graph (i.e., user dimension) and 1D graph Fourier transform. For example, this data can be the $i$-th column in Figure \ref{fig:interaction_matrix}(c), denoted as ${\bm{{\rm R}}}_{*i}$. The graph Fourier transform $\mathcal{F}_g({\bm{{\rm R}}}_{*i})$ is shown in Figure \ref{fig:intuitive_example}. Blue points indicate the useful signal and red points indicate the noise. Since noise is randomly distributed pulses, it is white noise and (approximately) keeps constant in the frequency domain. According to the property of Laplacian matrix, $\lambda_1=0$ \cite{hyper_laplacian}, thus the first blue point is the direct current (DC) component of ${\bm{{\rm R}}}_{*i}$. The DC component keeps constant among all users in the graph (time) domain hence indicates the popularity of item $i$ intuitively, i.e., the global preference towards $i$. Signal in the green rectangle is the low-frequency component of ${\bm{{\rm R}}}_{*i}$, which varies very smoothly among users and can be explained as the common preference of all users. The frequency of the signal in the yellow rectangle is higher, and this component varies more strongly in the graph domain, which can be explained as the specific (personalized) preference of users. Noise distributes on the high-frequency varies violently in the graph domain, which corrupt the signal thus should be removed by the low-pass collaborative filter. The low-pass filter is a gate function in the frequency domain (red line in Figure \ref{fig:intuitive_example}), which keeps the amplitude of low-frequency and reduces the amplitude of high-frequency to 0. We can see that by filtering, we remain the user preference and remove the noise.

\section{Illustration of LCFN model}
\label{app:illustration}
An illustration of LCFN is shown in Figure \ref{fig:GCN}. ${\bm{{\rm R}}}^{(0)}$ is the inputted matrix and $\{{\bm{{\rm R}}}^{(l)}\}_{l=1\cdots L}$ are feature maps. We fuse them by sum pooling and supervise the output by the observed data.

\section{Experiment Results}
\label{app:experiment_results}
The performances of recommending top-\{2, 5, 10, 20, 50, 100\} items are shown in Table \ref{tab:exper_result_all}. We discuss several interesting observations. First, comparing \textit{Amazon} and \textit{Movielens}, it is obvious that the dataset with higher sparsity (\textit{Amazon}) is more challenging to predict. However, LCFN gains more improvement on \textit{Amazon} than on \textit{Movielens}. The possible reason is that we face more serious sparsity issue on \textit{Amazon} dataset, where the observed data cannot provide enough information to supervise model training thus we benefit a lot from exploring the topology. 

Another observation also catches our interest is that the improvement LCFN gains reduces with the increasing of $k$. This may be because all models are tuned according to $F_1$-score@2, thus perform well for top-2 items recommendation. However, they are not well tuned for a large $k$ and advanced models fail to achieve the best improvement over the basic model. With the increasing of $k$, both LCFN and GNN/GCN baselines show less improvement over MF. To get better performance for a large $k$, we can retune all models according to $F_1$-score@100. 

\end{document}